# Multi-Objective Software Suite of Two-Dimensional Shape Descriptors for Object-Based Image Analysis

Andrea Baraldi and João V. B. Soares

*Abstract* — In recent years two sets of planar (two-dimensional, 2D) shape attributes, provided with an intuitive physical meaning, were proposed to the remote sensing community by, respectively, Nagao & Matsuyama and Shackelford & Davis in their seminal works on the increasingly popular geographic object-based image analysis (GEOBIA) paradigm. These two published sets of "simple" geometric features were selected as initial conditions by the present research and technological development software project, whose multi-objective goal was to accomplish: (i) a minimally dependent and maximally informative design (knowledge/information representation) of a general-purpose, user- and application-independent dictionary of 2D shape terms provided with a physical meaning intuitive to understand by human end users and (ii) an effective (accurate, scale-invariant, easy to use) and efficient implementation of 2D shape descriptors. To comply with the Quality Assurance Framework for Earth Observation (QA4EO) guidelines, the proposed suite of geometric functions is validated by means of a novel quantitative quality assurance ($Q^2A$) policy, centered on inter-feature dependence (causality) assessment. This innovative multivariate feature validation strategy is alternative to traditional feature selection procedures based on either inductive data learning classification accuracy estimation, which is inherently case-specific, or cross-correlation estimation, because statistical cross-correlation does not imply causation. The project deliverable is an original general-purpose software suite of seven validated off-the-shelf (ready-for-use) 2D shape descriptors intuitive to use. Alternative to existing commercial or open source software libraries of tens of planar shape functions whose informativeness remains unknown, it is eligible for use in (GE)OBIA systems in operating mode, expected to mimic human reasoning based on a convergence-of-evidence approach.

This work was supported in part by the National Aeronautics and Space Administration under Grant No. NNX07AV19G issued through the Earth Science Division of the Science Mission Directorate.

A. Baraldi was with the Department of Geographical Sciences, University of Maryland, College Park, MD 20742, USA. He is now with the Dept. of Agricultural and Food Sciences, University of Naples Federico II, Portici (NA), Italy (e-mail: andrea6311@gmail.com).
J. V. B. Soares is with the Department of Computer Science, University of Maryland, College Park, MD 20740 USA. e-mail: joao@cs.umd.edu.

## I. INTRODUCTION

Recent years have seen huge amounts of digital visual (two-dimensional, 2D, planar) data generated, transmitted, stored, retrieved and made accessible to a wide public of scientists and general users. Stemming from the ever increasing quality, quantity and accessibility of digital imagery, there is an urgent demand for effective and efficient computational vision tools, whose application domain ranges from low-level visual data representation and description to high-level image understanding (classification, i.e., three-dimensional (3D) scene reconstruction from (2D) imagery) [1], [2], [3], encompassing content-based image storage and retrieval (CBISR) system development [4], [5], [6], [7].

In human vision, 2D shape properties play a pivotal role [8], [9], [10], in combination with color [3], textural properties [3], [11], [12], [13], [14] and inter-object spatial relationships [15], [16], [17], [18]. When a 3D real-world object is projected onto a 2D image plane, one dimension of the 3D object information is lost. Due to dimensionality reduction [2], the 2D shape extracted from an image only partially represents the projected 3D object, i.e., 2D shapes can be affected by (self-)occlusion phenomena. In digital images, to make the 2D shape analysis and recognition problem even more complex, shapes of image-objects are also affected by spatial aliasing, due to the spatial resolution of the imaging sensor, and can be corrupted by photometric noise. In computational vision, effective and efficient solutions to the inherently difficult geometric problem of quantitative 2D shape analysis and recognition would impact many scientific domains [19], such as cognitive science, computer vision, which includes medical imaging and remote sensing (RS) imaging, computer aided design, molecular biology, geographic information systems, etc., see Fig. 1. For example, shape is one of the primary low-level image features investigated in existing CBISR systems, not yet available in operating mode to a general public [4], [5], [6], [7], [23].

(Fig. 1 about here)

(Fig. 2 about here)

In their survey of 2D shape descriptors developed by computer vision, computational geometry and computational morphology [9], [23], Zhang and Lu categorized as "simple" those planar object features provided with an intuitive physical meaning, see Fig. 2. Related to human perception, "simple" 2D shape descriptors are eligible for use in computer-based decision systems capable of mimicking human reasoning [24]. Computational geometry is a relatively new and flourishing discipline of computer science, coping with the systematic study of algorithms and data structures for 2D and 3D geometric problems [20], [25], [26], [27], [28], with a focus on exact algorithms that are asymptotically fast. Software projects that provide easy access to efficient and reliable 2D and 3D geometric algorithms and data structures (e.g., Delaunay tetrahedralization, Voronoi tessellation, surface and volume mesh generation, convex hull algorithms, 2D- and 3D-object skeletonization, etc.) in the form of a software library are the Computational Geometry Algorithms Library (CGAL) [27] and the Library of Efficient Data Types and Algorithms (LEDA) [28]. Unfortunately, software libraries such as CGAL and LEDA do not provide any suite of "simple" 2D shape descriptors.

Computational morphology is the study of form or structure as in the case of automatic recognition of shape by machines [26], [29]. In the context of morphological image analysis [29], software libraries of "simple" geometric attributes of planar objects are made available by, for example, the eCognition commercial software product [30], the Open Source Computer Vision Library (OpenCV) [31] (refer to Appendix 1) and the ENvironment for Visualizing Images (ENVI) EX 5.0 commercial software toolbox [32] (refer to Appendix 2). Consisting of around forty-five, fifteen and fourteen geometric descriptors respectively, these three collections of basic geometric terms are completely different from one another. This means they adopt different multivariate feature representation and description optimization criteria; above all, these optimization criteria are unknown to the scientific community to date. As a consequence, although employed on a regular basis by computer vision and RS scientists and practitioners, these suites of geometric functions are not provided with any known validation (*Val*) policy for quantitative quality assurance ($Q^2A$), to be community-agreed upon, as recommended by the Quality Assurance Framework for Earth Observation (QA4EO) guidelines [33] (refer to Appendix 3).

To contribute to filling the information gap from sensory "big data" to operational, comprehensive and timely information products subject to a *Val* policy for $Q^2A$ [33], the present interdisciplinary research and technological development (RTD) software project pursues an original multi-objective optimization of a dictionary of basic 2D shape terms. To make this inherently ill-posed software project better conditioned for numerical treatment, it is subject to the following constraints.

(1) Model design, regarded as "anything, but coding" [34]. The 2D shape representation consists of a discrete and finite set of quantitative variables, equivalent to a multivariate geometric indicator. It is expected to be: (i) general-purpose, i.e., data-, user- and application-independent [35], (ii) minimally redundant and (iii) maximally relevant [36], [37]. In the machine learning literature, these two latter conditions are known by acronym mRMR [36], [37]. For the sake of clarity and effectiveness, the mRMR criteria are reformulated in the present study. To account for the well-known dictum that "correlation does not imply causation", quantitative summary statistics of a complex target phenomenon are required to be minimally dependent and maximally informative, mDMI. The collective mDMI feature design optimization criteria are considered related to the problem solving principle known as Occam's razor[1] in the machine learning community [38], [39]. (iv) Made of "simple" perceptually significant 2D shape features. Provided with a physical meaning intuitive to understand [9], [23], these geometric features are eligible for use in the increasingly popular object-based image analysis (OBIA) paradigm, which includes geographic OBIA (GEOBIA) as a special case [40]. (GE)OBIA systems are expected to mimic human reasoning [1], [2], based on a convergence-of-evidence approach, where multiple fuzzy (eventually, weak) sources of converging perceptual evidence are combined to infer (eventually strong) conjectures [1], [2], in accordance with the general principles of fuzzy logic [41] and fuzzy decision trees [24], [42]. (v) Subject to a *Val* policy for Q$^2$A, in agreement with the QA4EO guidelines (refer to Appendix 3) [33].

(2) Model implementation. Each variable descriptor (extractor) is expected to be delivered in operating mode. To be considered operational (off-the-shelf, ready-for-use), each individual 2D shape descriptor must be optimized to jointly score "high" in a set of community-agreed quantitative quality indicators (Q$^2$Is) of operativeness (Q$^2$IOs) [15], [16], subject to a *Val* policy for Q$^2$A in compliance with the QA4EO guidelines (refer to Appendix 3) [33]. In the present work, for each geometric feature descriptor, adopted Q$^2$IOs are: (a) accuracy, (b) efficiency and (c) robustness to changes in the input dataset, including invariance with respect to translations, rotations and scaling transformations [15], [16], [43].

The aforementioned list of project requirements is different, either totally or in part, from those found in related works on geometric feature design and implementation, such as [8], [23], [35], [43], [44].

The original contribution of the present RTD software project is twofold. Validated by a mandatory Q$^2$A policy, the software project deliverable is a general-purpose suite of seven off-the-shelf 2D shape descriptors provided with an intuitive physical meaning. Pertaining to the fields of computational geometry [20], [25] and morphology [26], [29], "simple" 2D shape terms of certified quality can be integrated into software libraries of efficient and reliable computational geometry algorithms, such as CGAL [27]. Any general-purpose set of validated quantitative variables defined beforehand (equivalent to past knowledge) can be selected by a wide public of scientists and practitioners as a reliable initial condition (first choice), to be subject to further case-specific adjustments by means of data-, user- and application-dependent inductive data learning algorithms [39] for feature pruning [36], [37], feature mixture (e.g., principal component analysis [45]) or supervised data learning classification [38], [39]. For example, a popular statistical model for feature selection is stepwise regression. This is a greedy approach (globally suboptimal, but stepwise locally optimal) that adds the best feature (or deletes the worst feature) at each round according to an optimization criterion, such as mRMR [36], [37]. It is important to remind non-expert users that any inductive data learning algorithm is inherently ill-posed and requires *a priori* knowledge in addition to data to become better posed for numerical treatment [38], [39]. It means that no learning-from-examples algorithm (equivalent to phenotype in biological learning systems [46], [47]) should ever be confused with its initial conditions (equivalent to genotype in biological learning systems [46], [47]). In the words of genetic epistemology, for any (biological) cognitive system, "there is no absolute beginning" [46], [47]. By analogy, the application-independent dictionary of basic terms proposed in this study, equivalent to initial conditions driven from an *a priori* (deductive, top-down) knowledge base available in addition to data, should never be confused with any application-specific inductive (bottom-up) data learning algorithm, which explores the neighborhood of its initial conditions in a solution space based on evidence collected from the available dataset [47].

---

[1] A problem solver should always prefer a simpler model to more complex models and this preference should be traded off against the extent to which these alternative models fit the data.

The second original contribution of the present study is the proposed *Val* strategy for $Q^2A$ of a set of quantitative random variables, whose dependence (causality [48]) must be minimized (mD). It comprises a three-level decision tree where the Pearson's chi-square test for statistical independence [49], [50], [51], the Spearman's rank cross-correlation coefficient (SRCC) [52], [53] and a local proof of the absence of monotonically increasing or decreasing pairwise feature relationships are estimated hierarchically. This multivariate feature $Q^2A$ policy is a viable alternative to statistical techniques for feature design and/or selection adopted by the mainstream RS community, including: (i) maximization of a multivariate classification accuracy [54], [55]. Any inductive data learning classification algorithm is inherently case-specific [38], [39], [56] and ignores other important $Q^2$IOs of features descriptors, such as efficiency, whose relevance is high in other application domains, such as CBISR. (ii) Minimization of redundancy (mR), where redundancy is intended as inter-variable cross-correlation. Traditionally, cross-correlation is implemented by the Pearson's cross-correlation coefficient (PCC) [35], sensitive to statistical linear relationships (collinearities). It is well known that "correlation does not imply causation" [52], [53]. It is also true that causation that does not imply cross-correlation. For example, a non-linear inter-variable dependence (causal function) can cause PCC to be zero[2]. The hierarchical $Q^2A$ protocol for mD of a feature set proposed in this study can be applied to any multivariate feature analysis problem. For example, it could be adopted by software developers of the eCognition commercial software product [30], the OpenCV library [31] and the ENVI EX 5.0 commercial software toolbox [32], whose multi-objective optimization criteria for 2D shape index representation and implementation remain unknown, to comply with the *Val* requirements of the QA4EO guidelines.

The rest of this paper is organized as follows. Section II reviews the problem terminology. Related works are summarized in Section III. Materials and methods are described in Section IV and Section V respectively. Experimental results are presented in Section VI and discussed in Section VII. Conclusions are proposed in Section VIII. For comparison purposes and to make the paper self-contained, Appendix 1 and Appendix 2 present, respectively, the list of geometric functions implemented in OpenCV [31] and in the ENVI 5.0 commercial software toolbox [32]. Provided with a relevant survey value, Appendix 3 summarizes the *Cal/Val* requirements of the QA4EO guidelines [33].

## II. TERMINOLOGY

Our investigation focuses on 2D shape descriptors exclusively. In this context, to avoid possible ambiguities in terms, let us introduce some terminology first. According to the Open Geospatial Consortium (OGC) nomenclature [57], a segmentation map is defined as a plane (image) partition consisting of mutually exclusive and totally exhaustive *plane entities*, where each part is identified by an integer value. As a consequence, a segmentation map is a multi-level image [42], where the number of levels equals the number of plane entities. Planar entities, termed (2D) *geometric objects* or *geometric primitives* in the OGC dictionary [57], are typically called (2D) tokens, segments, regions, patches or image-objects in the RS and computer vision literature [40], [42]. In the OGC terminology [57], the base *Geometry* class of geometric objects has four subclasses: (0D) *Point* (representing a single location in coordinate space; the boundary of a Point is the empty set), (1D) *Curve* (an open planar curve [43], defined as a connected sequence of Points), (2D) *Surface* (associated with an "exterior boundary" and zero or more "interior boundaries") and *GeometricCollection*, which combines entities of the former subclasses. Examples of *atomic geometric types* of a geometric object, identified by an integer code, are: (0D) *Point* (code 1), (1D) *LineString* (code 2), (2D) *Polygon* (code 3), *MultiPoint* (code 4), *MultiLineString* (code 5), *MultiPolygon* (code 6, whose Polygon elements cannot intersect, but may touch), etc.

---

[2] For example, it is easy to prove that PCC(x,y) = cov(x,y) / $\sigma(x)\sigma(y)$ = (E[xy] − E[x]E[y]) / $\sigma(x)\sigma(y)$ is zero if y = x^2 with x in [-1, 1].

Spatial attributes of plane entities comprise positional and shape attributes, if any. In existing commercial or open source software libraries developed for computer vision and/or RS image processing (enhancement) and understanding applications, geometric attributes are extracted from image-objects whose atomic geometric type is either (1D) *LineString* (code 2) or (2D) *Polygon* (code 3), or, vice versa, from open or closed image contours. Noteworthy, image contour detection is the dual problem of image segmentation [2] and these are both inherently ill-posed problems [11] in the Hadamard sense [58], whose solution does not exist or is not unique or, if it exists, it is not robust to small changes in the input dataset.

## III. RELATED WORKS

The taxonomy of 2D shape features proposed by Zhang and Lu [9], [23] comprises three dichotomous levels of feature categories, summarized as follows, see Fig. 2.

- Spatial and transformed domain representations. Some of the former, called "simple" shape descriptors [35], hold an intuitive physical meaning, like convexity, compactness, elongatedness, etc., such that their behavior can be intuitively predicted [43]. Other geometric descriptors in the spatial domain are not intuitive to interpret, but have the desirable properties of being invariant under translation, scaling and rotation, like the either area- or contour-based Hu geometric moments [59]. Traditionally, geometric operators in the spatial domain suffer from two main drawbacks: noise sensitivity and high dimensionality. These problems can be solved by analyzing shape in a transformed domain, like the spectral domain. Examples of spectral descriptors are the Fourier descriptor (FD) and the wavelet descriptor (WD) [8], [9], [10].

- Contour-based and region-based representations. Among geometric operators in the spatial domain, traditionally considered affected by high sensitivity to noise and high dimensionality, area-based descriptors are considered more robust, i.e. less sensitive to noise or shape deformations, while boundary-based descriptors are considered more sensitive to changes in the input dataset [23], [43].

- Continuous (global) as opposed to discrete (structural) representations. In the former, a planar shape is represented as a whole, such that the resulting representation is a quantitative feature vector. In the resulting multi-dimensional shape space [11], different shapes correspond to different points in this space and a quantification of shapes differences is accomplished using a metric distance between the acquired feature vectors, e.g., Hamming distance, Hausdorff distance, comparing skeletons and support vector machine pattern matching [60]. The latter represents a shape by sections, or primitives. For example, a shape boundary can be broken down into line segments by polygon approximation [9], [23]. In this case, the similarity measure between two shapes can be estimated by string matching or graph matching.

Peura and Iivarinen advocated the use of semantically "simple" shape descriptors whenever possible [35]. Their heuristic multi-objective feature representation and description optimization criteria are verbally, rather than quantitatively, expressed as follows. The feature set should be: (i) simple, intended as compact. We interpret this expression as minimally redundant, mR, because Peura and Iivarinen also stated that "some correlation between descriptors is acceptable", where inter-feature (cross-)correlation was estimated as the PCC [51], [52], [53]. (ii) Generally applicable. To be general-purpose, it must be maximally relevant (MR), in any application domain. In qualitative terms, Peura and Iivarinen wrote that "combining descriptors should introduce a new perspective". Finally, individual descriptors should be: (iii) computationally efficient and (iv) intuitive to understand, i.e., "each descriptor should be semantically simple". For $Q^2A$ of a feature set of five "simple" shape descriptors, namely, convexity, ratio of principal axes, compactness, circular variance and elliptic variance, Peura and Iivarinen selected a test set of 79 planar shape instances and estimated the PCC value for each feature pair.

In agreement with Peura and Iivarinen [35], Zhang and Lu [9], [23] observed that semantically "simple" shape descriptors are not suitable as standalone descriptors, but a combination of descriptors is necessary in order to accurately describe shapes. This domain-specific statement can be regarded as common knowledge to cope with the *non-injective property of any summary (gross) statistic* or $Q^2I$, which implies that no "universal" $Q^2I$ can exist, because two different instantiations of the same target phenomenon can feature the same summary statistic [61]. Largely oversight in common practice, the non-injective property of $Q^2Is$ is inconsistent with a traditional search for universal image quality indexes (UIQIs), still ongoing by a relevant portion of the computer vision community [62], [63], [64], [65], [66]. By analogy, in economic studies, no economic univariate (scalar) $Q^2I$, such as the popular gross domestic product, should ever be considered *per se*, but in an mDMI combination with other $Q^2Is$ [36], [37], such as the Gini index estimating the inequality of wealth, a pollution/environmental quality index, etc. [67]. To conclude, due to the non-injective property of summary statistics and in agreement with common sense, any quantitative investigation of a target complex phenomenon through summary statistics should avoid univariate analysis of one "universal" (scalar) $Q^2I$, which cannot exist in practice, in favor of a multivariate variable analysis, where an mDMI dictionary of $Q^2Is$ must be designed and implemented in compliance with the Occam's razor principle [38], [39]. For *Val* purposes, a multivariate feature $Q^2A$ protocol for mDMI optimization can employ a multi-objective convergence-of-evidence approach [2], which is a key decision strategy in cognitive systems [24], [41], refer to Section I.

In the domain of computational morphology, the eCognition commercial software toolbox [30], the OpenCV library [31] (refer to Appendix 1) and the ENVI EX 5.0 commercial software product [32] (refer to Appendix 2) include very different ensembles of "simple" geometric descriptors (refer to Section I), whose atomic geometric type is either (1D) *LineString* (code 2) or (2D) *Polygon* (code 3, refer to Section II), or, vice versa, open or closed image contours. Unfortunately, none of these available libraries of basic geometric functions was subject to any known *Val* strategy for feature selection/design and extraction/implementation) $Q^2A$.

In the RS literature, two compact sets of "simple" planar geometric attributes were proposed by, respectively, Nagao & Matsuyama [1], [2] and Shackelford & Davis [68], [69], whose seminal works pioneered the increasingly popular GEOBIA paradigm [40]. Encompassing both contour-based methods and region-based algorithms in the spatial domain, see Fig. 2, these two known sets of intuitive geometric properties are jointly reviewed hereafter.

- Oriented minimum enclosing rectangle (MER), parameterized as width (W), length (L) and orientation angle ($\alpha$) of the L side, with L $\geq$ W.
- Size, parameterized as area (A) in pixel unit.
- Elongatedness (*Elngtdnss*) $\geq$ 1, estimated with several approaches, refer to [1], [2], [68] and [69] for further details.
- Compactness (*Cmpctns*) or circularity, estimated from the planar object's area, A, and perimeter, P. Traditionally, it is estimated in a variety of different formulations [1], [35], [70], [71], [72], [73], e.g., *Cmpctns* = 2sqrt(A$\pi$)/P $\in$ [0, 1] [35]. To increase confusion, some authors deal with noncompactness, e.g., $P^2$/A $\geq$ 1 [1], [70], [73]. Most of these formulations are not scale invariant.
- Single-resolution (vice versa, single-scale) straightness of boundaries [1], [2], as an indicator of manmade Earth surface objects observed from space. This geometric feature is absent from open source and commercial software libraries of planar geometric functions, such as [30], [32] and [57].
- List of the object's vertices after polygon skeletonization, where vertex attributes are position, angle and inter-endpoint distance [68], [69].
- Fuzzy rectangularity, particularly useful in RS images for building detection. An "approximately rectangular" image-object model of real-world buildings is defined as a 2D shape satisfying three properties: (i) about four endpoints with angles close to 90° and "large" separation (defined in this case larger than 5 m), (ii) about two or less endpoints with angles much larger than 90° and (iii) about two or less endpoints with angle much smaller than 90°. Because the attributes of an "approximately rectangular" image-object are imprecise and the shapes of buildings vary, fuzzy membership functions are adopted to measure how closely the endpoint angles and the line segment lengths between the endpoints match different physical model-based decision rules [68], [69].

Noteworthy, neither Nagao & Matsuyama [1], [2] nor Shackelford & Davis [68], [69] discussed their quantitative multi-objective optimization criteria for geometric feature design and implementation.

In the Moving Pictures Expert Group (MPEG)-7 Visual Standard proposed to search, identify, filter and browse audiovisual content, several principles to measure a shape descriptor have been set: good classification/retrieval accuracy, compact features, general application domain, low computation complexity, retrieval performance robust to noisy data and hierarchical coarse-to-fine representation [8], [9], [10]. In MPEG-7, the either area- or contour-based multiscale generic Fourier descriptor (GFD) was considered to be the single best choice to accomplish the six principles set by MPEG-7 [8], [9], [10]. Although GFD has a physical interpretation and is particularly effective for quantitative shape matching and retrieval, it is not provided with an intuitive meaning. Hence, it is not suitable for implementation in computer vision systems expected to mimic human reasoning based on intuitive decision rules, such as GEOBIA systems [1], [2], [68], [69].

## IV. Materials

To comply with standard evaluation criteria requiring a minimum of two real or realistic test datasets [74], three populations of geometric polygons were selected for testing purposes.

(i) One synthetic dataset of around thirty 2D shapes, shown in Fig. 4, provided an environment of controlled complexity suitable for human decision makers (HDMs) to test whether geometric operators satisfy quantitative theoretical expectations and qualitative human perception.

(ii) A first real-world dataset consisting of around 8000 geometric objects automatically detected in a very high resolution (VHR) satellite optical image by the Satellite Image Automatic Mapper™ (SIAM™), a prior knowledge-based vector quantizer capable of automatic MS image preliminary classification (pre-classification) and segmentation, proposed to the RS community in recent years [75], [76], [77]. Image-objects detected by SIAM are spectrally uniform, equivalent to textons (texture elements) detected at the raw primal sketch in low-level vision [3]. Spectrally uniform VHR image-objects are typically affected by irregular shapes and by the presence of inner holes. It means that, although all of these 2D segments are individually connected, only some of them are simply connected [75], [77].

(iii) A second real-world dataset consisting of around 300 images of individual leaves, acquired against a white background by a consumer-level digital camera, was automatically segmented into a binary map according to the near real-time segmentation algorithm proposed in [78], [79].

No segmentation quality assessment was pursued in this paper, since this task would go far beyond the goal of the present RTD project, reported in Section I. In practice, segments were considered equivalent to an *a priori* knowledge which is, by definition, available beforehand in addition to data (observables, true facts).

## V. METHODS

In line with the words by Piaget (in cognitive systems "there is never an absolute beginning" [46], [47]), the proposed RTD software project moved from the legacy of past works by Nagao & Matsuyama [1], [2] and Shackelford & Davis [68], [69], summarized in Section III and selected as initial conditions, to design and implement an original software dictionary of "simple" 2D shape descriptors, where the atomic type of target geometric objects is either (1D) LineString (code 2) or (2D) Polygon (code 3), refer to Section II, subject to multiple objectives to be jointly optimized in agreement with the Occam's razor, refer to Section I.

A formal analysis of multi-objective problems, where many possible courses of action are competing for attention, is due to the Italian civil engineer, economist and sociologist Vilfredo Pareto [80]. In his terminology, given a set of $Q^2$Is to be maximized, called Pareto dimensions, the so-called Pareto efficient frontier (PEF) is the set of choices that are Pareto efficient [80]. A multi-objective solution is called non-dominated, non-inferior, Pareto optimal or Pareto efficient if none of the objective functions can be improved in value without degrading some of the other objective values. By restricting attention to the set of choices that are Pareto efficient, an HDM can make subjective tradeoffs within the set of Pareto-efficient solutions, rather than considering the full range of every parameter. Without additional preference information by an HDM, which means without subjective (equivocal, qualitative) prior knowledge available in addition to quantitative (objective, unequivocal) data, all Pareto-efficient solutions must be considered equally good, i.e., in general, a multi-objective optimization problem is inherently ill-posed in the Hadamard sense [58]. When decision making is emphasized, i.e., when one single model solution must be chosen, an HDM has to select the most preferred Pareto optimal solution along the PEF according to his/her own preference (*information-as-data-interpretation*, also refer to Appendix 3).

Belonging to the class of inherently ill-posed multi-objective optimization problems subject to the Pareto's formal analysis, our software design and implementation project is expected to admit no single "best" solution based on quantitative data analysis (*information-as-thing*, also refer to Appendix 3) exclusively.

*A. Qualitative Maximization of Informativeness of a Feature Set*

To maximize informativeness (MI) of a feature set in a general-purpose application domain as required by Section I, Peura and Iivarinen wished that "combining descriptors should introduce a new perspective". To our best knowledge, no quantitative case-independent feature design/selection strategy capable of accomplishing the MI constraint can be found in existing literature [8], [23], [35], [43], [44]. In computational morphology, we may distinguish between two phases of a shape detection process [26]. First, a multivariate shape parameterization (analysis) occurs, where a finite and discrete set of geometric features is selected and extracted. Second, any quantitative instantiation of a multivariate shape variable is mapped onto a discrete and finite set of mutually exclusive and totally exhaustive classes [81]. Typically, a multivariate data classification/retrieval accuracy index is adopted as an "overall" estimate of the effectiveness of the geometric feature representation and description sequence. However, as reported in Section I, any inductive data learning classifier is inherently case-sensitive [35], [56]. It means that, to validate the MI property of a feature set in terms of classification accuracy on a general-purpose basis, a large variety of inductive classification algorithm implementations should be tested by many different users upon a large ensemble of different input datasets. In addition, the effectiveness (accuracy) of features employed to retrieve similar shapes from a designated database is not sufficient to guarantee the $Q^2A$ of a shape selection and representation approach, because other $Q^2IOs$, such as efficiency, can be important in other application domains, such as CBISR [23].

To explain the ongoing lack of a standard community-agreed quantitative assessment of the MI property of a feature set in an application-independent scenario, we observe that the degree of informativeness of a set of features is an inherently equivocal (subjective, qualitative) cognitive problem (*information-as-data-interpretation*). As such, it does not pertain to the quantitative unequivocal domain of *information-as-thing* [4], refer to Appendix 3. If this consideration holds true, then this paper can provide no realistic $Q^2I$ value to answer questions like: in general, does *Roundness* bear some useful information in addition to *Convexity*? Can *Rectangularity* be considered useful in the computer vision and RS common practice? Does feature *Straightness-of-boundaries*, proposed by Nagao & Matsuyama, but not implemented by any open source or commercial software library, such as [30], [32] and [57], "introduce any new perspective" [35]? These difficult (ill-posed) questions admit more than one (subjective) solution. Because estimating the collective degree of informativeness of a feature dictionary in a general-purpose application domain is an inherently equivocal *information-as-data-interpretation* problem [4], our realistic conclusion is that only "relative" (qualitative, subjective, equivocal) decisions, rather than "absolute" (quantitative, objective, unequivocal) decisions can be drawn by an HDM, based on his/her own preference or past knowledge, independent of and in addition to available observables (true facts), if any [38], [39].

*B. Quantitative Minimization of Dependence of a Feature Set*

To account for the well-known dictum that "correlation does not imply causation", the present work replaces the traditional multivariate feature optimization criterion of minimization of redundancy (mR), where redundancy is estimated as the PCC, which is equivalent to collinearity, with the criterion of minimizing an inter-feature degree of dependence (mD), where dependence means causality, refer to Section I. The multivariate feature mD criterion can employ a goodness-of-fit (GOF) test well known in statistics, such as the Pearson's chi-square test of independence between two categorical random variables [49], [50]. Unlike inherently ill-posed inductive data learning algorithms, which require prior knowledge in addition to data to become better posed for numerical treatment [38], [39], GOF tests employ no prior knowledge, but a reference model of a statistical distribution. On the one hand, GOF tests are, so to speak, "more inductive" (data-driven) and "less deductive" (prior knowledge-based, to become better posed) than inductive data learning algorithms for feature selection and data classification. On the other hand, due to the inherent subjectivity of any *information-as-data-interpretation* process (refer to Appendix 3), GOF test results, just like outcomes collected from any inductive data learning algorithm, must be carefully scrutinized for interpretation by human experts, as clearly acknowledged by statisticians in the following quote [60].

> "Care must be taken not to over-interpret or over-rely on the findings of goodness-of-fit (GOF) tests. It is far too tempting … to run GOF tests against a generous list of candidate distributions, pick the distribution with the 'best' goodness-of-fit statistic, and claim that the distribution that fit 'best' was not rejected at some specific level of significance. This practice is statistically incorrect and should be avoided. GOF tests have notoriously low power and are generally best for rejecting poor distribution fits rather than for identifying good fits… GOF tests may, at best, simply serve to confirm what the analyst has found through visual inspection of the probability plots and other comparisons … The human eye and brain are able to interpret and understand data anomalies far beyond the ability of any computer program or GOF test" [60].

To implement a multivariate feature mD criterion, we started from the formal analysis of causal models proposed by Pearl [48]. A causal model is a directed acyclic graph G, meaning a (directed) graph which has no loops in it. In a causal graph G, each vertex (node) has an associated random variable, all root vertices (i.e., vertices with no parents) are labelled by independent random variables [48] and each oriented arrow indicates the possibility of a direct causal influence between two random variables. In a causal graph G, causality means that for any particular variable $X_j$, the only random variables which directly influence the value of $X_j$ are the parents of $X_j$, i.e., the collection of nodes $X_{pa(j)}$ of random variables which are connected directly to $X_j$ [48]. If two random variables X and Y, computed as deterministic functions (features) of an individual entity E (e.g., a plane entity) belonging to a sampled population P of entities, such that $E \in P$, where $X = F1(E)$ and $Y = F2(E)$, are non-causal, i.e., X does not belong to set pa(Y) and Y does not belong to set pa(X), then it would be impossible to indicate which variable predicts or causes the other, see Fig. 3.

To reduce the risk that, in a target set of random variables, a causal relationship exists between any possible pair of random variables we adopted a hierarchical combination of sufficient but not necessary conditions for independence, collected at different levels of detail, from "global" statistics (properties of a population P of entities Es) to "local" properties of individual entities, Es. The rationale of the adopted convergence-of-evidence approach is discussed below.

(Fig. 3 about here)

*1) First sufficient not necessary criterion for pairwise feature non-causality: Statistical independence*

The popular PCC, traditionally adopted to investigate pairwise feature redundancy considered equivalent to collinearity  [35], [62], [63], [64], [65], [66] (refer to Section I), is replaced by the Pearson's chi-square test for random variable independence, well known in statistics [49], [50], [51], [52], [53]. By definition, two categorical random variables X and Y are statistically independent if the occurrence of one does not affect the probability of the other, i.e., $p(X \cdot Y) = p(X) \cdot p(Y)$. It means that if two variables are statistically dependent, then knowledge of the level (category) of X *can* help predicting the level of variable Y, or vice versa. In other words, if two variables are statistically independent, it is impossible to indicate which variable predicts or causes the other. If there is a causal function of any degree between two random variables, then they are statistically dependent, although their PCC value can be zero, refer to footnote 2. On the other hand, the vice versa does not hold, i.e., if two variables are statistically dependent, they are not necessarily linked by a causal relationship. For example, let us consider two "simple" shape descriptors, such as *Compactness (Roundness)* and *Convexity*. Based on object-pair examples at a "local" scale of analysis (irrespective of "global" trends), it is clear that if *Roundness* is high then *Convexity* is also high. When *Roundness* is low, *Convexity* can be either high or low. Vice versa, when *Convexity* is high then *Roundness* can be either high or low. When *Convexity* is low, *Roundness* is also low. Hence, given a generic population of geometric objects, variables *Roundness* and *Convexity* are expected to perform as dependent variables to some (high) degree, because when the level of *Roundness* is known, it *can* help predicting the level of *Convexity* and vice versa. But (high) dependence does not mean there is a causal relationship between the two variables. To summarize, statistical independence is a sufficient, but not necessary condition for non-causality. If statistical dependence is "high", then causality does not necessarily holds. When the Pearson's chi-square test for independence is adopted to accomplish the mD feature constraint, the traditional project requirement specification that "some correlation between descriptors is tolerated" [35] becomes: "some statistical dependence between descriptors is tolerated", if further investigations are conducted to avoid the risk of inter-variable causality. As such, statistical (in)dependence is far more informative (useful) than PCC for investigating inter-variable causality.

In the RS and computer vision common practice, the reason why the PCC is traditionally adopted for bivariate feature redundancy assessment, while the Pearson's chi-square test for independence is almost never investigated, is that the former is input with two quantitative variables, which are very common in sensory data applications, while the latter requires as input two qualitative (categorical, nominal) variables [49], [50]. To transform a quantitative variable into a nominal one, it must be quantized (discretized) into a finite and discrete set of k levels (buckets, bins, categories, intervals), at the cost of a superior degree of quantitative data pre-processing. To provide a reasonable partition of the range of chance of the quantitative variable at hand while accounting for the variable distribution, these k buckets are recommended to be equiprobable and their total number is selected based on heuristic criteria [82]. A popular choice is:

$$k = 2*N^{(2/5)}, \tag{1}$$

where N is the finite size of the sample dataset [82].

The proposed statistical test for independence consists of four steps [49]: (1) state the hypotheses, (2) formulate an analysis plan, (3) analyze sample data, refer to further Section VI, and (4) interpret results, refer to further Section VII.

(1) State the null and alternative hypotheses. Null hypothesis, H0: Categorical variables X, featuring R levels (like rows), and Y, featuring C levels (like columns), are independent. The alternative hypothesis, Ha, is that knowing the level of categorical variable X can help predicting the level of categorical variable Y and/or vice versa. If this inter-variable relationship (dependence) holds, it is not necessarily causal.

(2) Statistical analysis plan. To describe how to use sample data to accept or reject the null hypothesis H0, the statistical plan specifies:

   a) The significance level $\alpha$, such that the degree of confidence is (1 - $\alpha$) [83]. Typically $\alpha$ is selected equal to the probability value 0.05, hence the degree of confidence is 0.95. There is one critical value (CV) of the target statistical distribution that corresponds to a given significance level. The probability that values of the statistic distribution are equal or superior to the CV is equal to the significance level.

   b) The test method. The well-known Pearson's chi-square test for independence is adopted. To be considered appropriate, it requires the following conditions to be met [49], [50]: (i) the sampling method is simple random sampling, (ii) each population is at least 10 times as large as its respective sample, (iii) the variables under study are both categorical, and (iv) the expected frequency count for each cell of the two-way contingency table is at least 5.

(3) Analyze sample data. Applying the Pearson's chi-square test for independence to sample data requires to compute the degrees of freedom, the expected frequency counts and the chi-square test statistic output value. The P-value is defined as the probability of observing a sample statistic value equal or superior (i.e., not inferior) to the obtained test statistic value.

(4) Interpret result. If the P-value is below the chosen level of significance $\alpha$, then H0 is dismissed at the level of confidence $(1 - \alpha)$. Similarly, given some significance level $\alpha$ and its corresponding CV, if the test statistic value is higher than the CV, then H0 is dismissed at the level of confidence $(1 - \alpha)$, i.e., there is an $\alpha$ probability that H0 is rejected by mistake [83] (refer to further Section VII).

For statistical data analysis, the popular Windows Excel software toolbox was adopted to quantize a quantitative random variable into k equiprobable bins, with k defined by Equation (1). Since the test statistic is a chi-square, we use the Windows Excel CHITEST() function to assess the P-value associated with the estimated chi-square test statistic value at the known value of degrees of freedom and significance level (refer to further Section VI).

*2) Second sufficient not necessary criterion for pairwise feature non-causality: Inter-feature statistical non-monotonicity (at the population level)*

If the statistical dependence between two random variables is "high", then we hierarchically consider a sufficient, although not necessary, condition for two geometric variables $X = F1(P)$ and $Y = F2(P)$, generated from the same population P, to be non-dependent, such that neither function $X = F3(Y)$ nor function $Y = F4(X)$ exists, see Fig. 3, if those two random variables X and Y are neither monotonically increasing nor decreasing. In statistics, the SRCC is a widely adopted nonparametric measure of statistical dependence between two ranked variables, refer to Section I. In greater detail, SRCC $\in$ [-1, 1] assesses how well the relationship between two ranked quantitative variables can be described by a monotonically increasing or decreasing function. If there are no repeated data values, a perfect SRCC of +1 or −1 occurs when each of the variables is a monotonically increasing or decreasing function of the other, even if their relationship is not linear, which makes SRCC quite different from PCC [51]. Traditionally, a cross-correlation coefficient greater than 0.80 represents strong agreement, between 0.40 and 0.80 describes moderate agreement, and below 0.40 represents poor agreement [81]. To summarize, when the statistical dependence between two random variables is "high", then we require their SRCC value to be "not high", e.g., SRCC < 0.8, to reduce the risk of inter-feature causality.

In our experiments, the popular Windows Excel software toolbox was adopted to generate ranked versions of two random variables to be input to the SRCC formulation.

*3) Third sufficient not necessary criterion for pairwise feature non-causality: Inter-feature local non-monotonicity at the individual level*

Last but not least, we must take into account the typical spatial nonstationarity of image statistics. In an image, "a localized image quality measurement can provide a spatially varying quality map of the image, which delivers more information about the quality of the image and may be useful in some applications" ([62], p. 7). It is like saying that (local) counterexamples are known to be helpful because they quickly show that certain (general) conjectures, or ideas or theorems, are false. This allows scientists to save time and focus their efforts on ideas to produce provable theorems.

To try to avoid inter-feature causality, if the statistical dependence between two random variables X and Y is "high" at global (image-wide) scale and their SRCC value is also "high" at global scale, then we require to find "local" evidence that the relationship between the two random variables X and Y cannot be "locally" described by a monotonically increasing or decreasing function. Hence, if the statistical dependence between two planar shape features is "high", but there is either global or local evidence that those two shape features are neither monotonically increasing nor monotonically decreasing, then those two features cannot be considered affected by a causal relationship, i.e., they cannot be considered the same feature, because it would be impossible to indicate which variable predicts or causes the other. In practice, the "local" absence of a monotonically increasing or decreasing function between two shape features X = F1(E), e.g., roundness, and Y = F2(E), e.g., convexity, of a planar entity E is verified if, in the target population of planar objects, at least one 3-tuple of entities, identified as R, E1 and E2, can be found such that, with respect to the reference instance, R, either one of the following two conditions (A) and (B) holds true.

Condition (A):
{ [(F1(E1) > F1(R)) AND (F2(E1) ≥ F2(R))]
  AND
  {
    [(F1(E2) ≥ F1(R)) AND (F2(E2) < F2(R))]
    OR
    [(F1(E2) ≤ F1(R)) AND (F2(E2) > F2(R))]
  }
}
OR Condition (B):
{ [(F1(E1) < F1(R)) AND (F2(E1) ≤ F2(R))]
  AND
  {
    [(F1(E2) ≤ F1(R)) AND (F2(E2) > F2(R))]
    OR
    [(F1(E2) > F1(R)) AND (F2(E2) ≤ F2(R))]
  }
}

## VI. Experimental Results

The objective of the experimental session is twofold. First, to comply with Section V, the experimental session considers a pair of 2D geometric descriptors as minimally dependent (mD) if there is no multi-level evidence of their causal relationship. This occurs if: (i) their statistical dependence is not "high" (the null hypothesis, H0, is accepted, refer to Section V.B.1), or (ii) their statistical dependence is "high" (H0 is rejected), but their SRCC value is not "high" (< 0.8, refer to Section V.B.2), or (iii) their statistical dependence is "high" (H0 is rejected), their SRCC value is "high" (≥ 0.8), but there is local evidence they are neither monotonically increasing nor decreasing (at any degree of relationship, whether linear or not), refer to Section V.B.3.

Second, to comply with Section I, the experimental session selects, for any proposed geometric attribute, the "best" descriptor, among possible alternative algorithms and implementations, that jointly maximizes the selected battery of $Q^2$IOs, consisting of: (I) invariance with respect to translations, rotations and scaling transformations, (II) robustness to noisy data, (III) minimization of computation time and memory occupation.

### A. *Original Representation and Implementation of 2D Shape Features whose Q²IOs Must Score High*

Inspired by the two sets of 2D shape descriptors proposed by Nagao & Matsuyama [1], [2] and Shackelford & Davis [68], [69], summarized in Section III, the following original set of contour-based 2D shape attributes is proposed in the spatial domain.

- Angle (orientation) of the segment's minimum enclosing rectangle [1], [84], not further discussed.
- Convexity, which is decreasing with the presence of holes (*CnvxtyAndNoHole*) $\in [0, 1]$, measured from the convex hull estimated from the outer boundary [42], refer to Section VI.A.1. It is absent from the two sets of attributes proposed in [1], [2] and [68], [69].
- Fuzzy rule-based rectangularity (*FuzzyRuleBsdRctnglrty*) $\in [0, 1]$, refer to Section VI.A.3. It employs a polygonal 2D shape representation alternative to the skeletonization algorithm adopted in [68], [69], refer to Section VI.A.2.
- Multiscale straightness of boundaries (*MltSclStrghtnsOfBndrs*) $\in [0, 1]$, alternative to the single-scale formulation proposed in [1]. Refer to Section VI.A.5.

Typically considered more robust to changes in the input dataset than contour-based attributes, area-based 2D shape attributes are estimated in the spatial domain according to the following original combination of features.

- Area (size), in pixel unit, not further discussed.
- Average per-segment characteristic scale (*DMPmltSclChrctrstc*) $\geq 1$, computed from the pixel-based morphological multiscale characteristic of the differential morphological profile (DMP) [85]. This characteristic scale value is proposed as a local spatial autocorrelation estimation, belonging to the class of local indicators of spatial association (LISA) [86]. This geometric feature is absent from the two sets of attributes proposed in [1], [2] and [68], [69]. See Fig. 4 and refer to Section VI.A.6.

(Fig. 4 about here)

(Fig. 5 about here)

- Scale invariant roundness (compactness, circularity), decreasing with the presence of holes (*RndnssAndNoHole*) $\in [0, 1]$. An original scale invariant compactness formulation is *RndnssAndNoHole* $= ((4 \times \mathrm{sqrt}(A)) / \mathrm{PL}) \in [0, 1]$, where PL is the 4-adjacency cross-aura measure estimated from the total perimeter, defined as the outer perimeter plus inner perimeter of holes, if any, see Fig. 5. To our best knowledge, this compactness equation is different from alternative formulations found in [1], [2], [35], [70], [71], [72]. Refer to Section VI.A.4.
- Elongatedness, increasing with the presence of holes (*ElngtdnssAndNoHole*) $\geq 1$. A new measure of *ElngtdnssAndNoHole* is presented to overcome operational limitations of the three formulations proposed by Nagao & Matsuyama, each at an increasing level of sophistication [1]. Refer to Section VI.A.7.
- *(Combined) Simple connectivity*, decreasing with the presence of holes (*CombndSmplCnctvty*) $\in [0, 1]$. This geometric feature is absent from the two sets of attributes proposed in [1], [2] and [68], [69]. It accounts for the dependence of region-based shape attributes on inner holes, whose presence may be an indicator of image noise. Refer to Section VI.A.8.

Noteworthy, planar shape operators belonging to range $[0, 1]$ can be byte-coded, with a quantization error of $(1. / 255) / 2 = 0.2\%$ when rounding to the nearest integer is adopted. For these operators, memory occupation is minimized.

Overall, seven 2D shape features are proposed, excluding area and orientation of the minimum enclosing rectangle, which are taken for granted. Hence, there are $\binom{7}{2}$ = 7! / 2! (7-2)! = 21 pairwise feature combinations to test for independence. The seven proposed descriptors and their related implementation issues are described hereafter.

*1) Area-based convexity sensitive to within-segment holes*

A commonly used shape property is convexity, also known as convexity ratio or solidity. It is defined as follows:

$$CnvxtyAndNoHole = A / Aconvex, \tag{2}$$

where the area value, A, is defined as the number of pixels that belong to the region, excluding those belonging to holes, if any, while *Aconvex* is the area of the convex hull of the region, which by definition ignores (includes) inner holes. Since inequality A ≤ *Aconvex* always holds true, then *CnvxtyAndNoHole* ∈ [0, 1]. The name of index *CnvxtyAndNoHole* is chosen to remind potential users that it takes on "high" values when the original segment is convex or close to being convex and, at the same time, it does not present holes. The first step in computing the area of the convex hull (*Aconvex*) is to find a representation of the hull, e.g., refer to the solutions proposed by the CGAL [27]. To detect the convex hull, we began by tracing the outer boundary of the original region using the standard boundary tracing algorithm attributed to Moore [87], [88]. This results in a sequence of pixels describing a walk along the region's boundary. From this sequence, the subset of the boundary pixels that form the set of vertices of the region's convex hull was selected using the algorithm independently discovered by Melkman [89] and Tor and Middleditch [90], which runs in O(n) time, with n equal to the original number of boundary points.

After finding the vertices of the convex hull, the area *Aconvex* can be computed from the vertices using the algebraic surveyor's (or shoelace) formula [91]. In digital imagery, this algebraic approach can lead to undesirable values of the *CnvxtyAndNoHole* variable. For example, a direct algebraic calculation of variable *Aconvex* for a region consisting of a straight line would be 0, whereas variable A would be the number of pixels in the line. Rather than trying to adapt the surveyor's formula to cope with 2D signal aliasing effects, we decided to estimate *Aconvex* by means of a discretization procedure, where a binary discrete image is overlapped with the algebraic convex hull, such that discrete pixels are set to 1 if the major portion of their area overlaps with the convex hull and 0 otherwise. *Aconvex* is then estimated as the number of pixels whose values is equal to 1.

(Fig. 6 about here)

*2) Polygonal representation of a 2D shape alternative to skeletonization*

Shackelford & Davis adopted a simplified polygonal representation of image-objects, to which a set of fuzzy logic rules was applied to decide about their shapes, e.g., approximate rectangularity, refer to Section III [68]. They obtained the polygonal representation of a geometric object from the endpoints of the region's skeleton, computed via a morphological thinning skeletonization, e.g., refer to the solutions proposed by the CGAL [27]. In our tests, this approach turned out to be problematic when the endpoints do not give a good description of an object's boundary, see Fig. 6. Inaccuracy of the endpoint-based description arose independently of the adopted skeletonization algorithm. To avoid this problem, a different strategy was implemented. It is common to approximate a contour representing a polygon with another contour having fewer vertices. This simplified polygon representation is typically obtained by means of the Ramer-Douglas-Peucker (RDP) approximation [92], [93], e.g., refer to the routine cvApproxPoly() in the OpenCV software library in Appendix 1 [31]. In addition to giving a better approximation of the region's boundary, this algorithm is also faster to run than skeletonization. At each step of the RDP algorithm, two points are considered. The points determine a line segment, which can be thought of as a rough representation of some part of the polygon. The point in the original polygon which is the farthest from this line segment and which lies in between the two points that determine the segment is then found. If the distance from this point to the line segment is smaller than a tolerance value ε, then the line segment is accepted as is. Otherwise, the problematic point is added to the representation, generating two new line segments which are recursively analyzed using the same procedure. We set the approximation tolerance ε equal to the scale corresponding to the maximum value of *Straightness$_s$*, refer to further Section VI.A.5. In other words, ε = argmax$_s$ *Straightness$_s$*, with s ∈ {1, 2, 4, 8, 16, 32}, where the observation scale parameter *s* in dyadic pixel units was used to compute the *Straightness$_s$* measure. Fig. 6 shows toy regions for which the skeleton-based polygon does not provide a good representation, whereas simplification with the RDP algorithm does.

*3) Contour-based fuzzy rectangularity*

Several measures of a geometric object's rectangularity have been proposed in the RS and computer vision literature. One standard measure is the area of the region relative to that of its oriented minimum enclosing rectangle [1]. Rosin further improved this measure by developing robust rectangle fitting procedures [42]. Additionally, he also presented a measure of rectangularity defined using the difference in moments between the region and its best fitting rectangle. In place of a traditional area-based rectangularity index, Shackelford & Davis defined a measure of "approximate rectangularity"' using a set of fuzzy rules [66]. These rules were applied to a polygonal representation obtained from a region's skeleton, as reviewed in Section VI.A.2. By using fuzzy rules to specify the expected number of vertices of a rectangle, their angles, and relative distance, the rectangularity measure becomes robust to a series of common shape variations, whether or not due to image noise. In the present study, the fuzzy rule-based rectangularity measure of Shackelford & Davis was selected due to its capability of modelling the within-class variance of rectangular shapes. Its implementation, identified as variable *FuzzyRuleBsdRctnglrty* ∈ [0, 1], was subject to two important changes. First, the polygonal approximation (simplification) of the geometric object at hand to which the rule set is applied was not obtained by area-based skeletonization, but with the contour-based RDP algorithm [90], described in Section VI.A.2. Second, to cope with noisy shapes and/or highly irregular segment boundaries, several of the Shackelford & Davis fuzzy rules' free-parameters were relaxed. In practice, the same set of fuzzy rules, consisting of S-, Z- and Π-membership functions [91], proposed by Shackelford & Davis was adopted [66], [67]. However, their implementation was relaxed to become less strict, which produced results more in line with human photointerpretation of 2D shapes affected by noise. For example, in the original fuzzy rule set proposed by Shackelford & Davis, to determine if a within-segment angle is around 90°, a Π-membership function is implemented with a so-called bandwidth, equal to the difference between the minimum and maximum acceptable values, set to 80° and 100° respectively. In our implementation suitable for noisy data, these two minimum and maximum parameters were relaxed to 60° and 120°. To summarize, our new implementation is more robust to changes in input data and computationally more efficient than its original counterpart by Shackelford & Davis.

*4) Area-based roundness (compactness) sensitive to within-segment holes*

An original *RndnssAndNoHole* (compactness) index formulation is proposed to be scale invariant and computationally efficient. A popular measure of a region's compactness, also called circularity or complexity, is $A/P^2$, with A denoting the region's area and P its perimeter [70]. The measure takes its maximum value of $1/4\pi$ when the region in question is a circle without inner holes. This consideration motivates the definition of a measure of *Roundness*, or circularity, as $4\pi A/P^2$, which always lies in the range [0, 1] [35]. Additionally, some authors adopt a measure of noncompactness, e.g., $P^2/A$ [1], [70], [73]. In general, a segment's area, A, is defined as the number of pixels in the region, where pixels belonging to holes, if any, are excluded from the estimation. However, in common practice, compactness is estimated after holes have been filled in, i.e., it is assumed the region is simply connected. A decision must be made on how to compute the perimeter P, since different definitions of a region's perimeter exist when dealing with digital images [42], [73]. In our original formulation of geometric index compactness, the perimeter length PL is the computationally efficient 4-adjacency cross-aura measure (see Fig. 5 [61]) of the region's total boundary, where the total boundary takes into account contributions from holes, i.e., total boundary = external (outer) boundary + inner boundary (due to holes) [42]. Starting from these definitions of area, A, and perimeter length, PL, the proposed formulation of *Roundness* becomes

$$RndnssAndNoHole = (4 \times \mathrm{sqrt}(A) \,/\, PL) \in [0, 1]. \qquad (3)$$

Reflected in the way both A and PL are computed, this measure treats holes as intrinsic properties of the geometric object, rather than filling them up. It scores high for regions that are round and, at the same time, do not have holes. This behavior justifies its name, *RndnssAndNoHole*. It can be easily proved (by induction) that index *RndnssAndNoHole* is scale invariant. For example, for an isolated 1-pixel object, PL = 4, then *RndnssAndNoHole* = 4/4 = 1 (maximum). For a 4-pixel square object, PL = 8, then *RndnssAndNoHole* = 4*2/8 = 1, etc. Although measures of compactness or roundness should be maximum for circles, Rosenfeld pointed out that, in digital images, depending on how the perimeter is measured, compactness measures could turn out to be larger for squares or octagons than for digitized circles [73]. Our measure is no exception and takes maximum values for squares, which have maximum area for a given fixed value of the 4-adjacency cross-aura measure.

*5) Contour-based multiscale straightness of boundaries*

A measure of the straightness of a region's outer boundary is especially discriminative for the analysis of RS images of buildings in urban areas or crop fields in agricultural land. In general, manmade Earth surface structures tend to present straight boundaries, independently of whether their overall shape is simple or more complex. Following the procedure proposed by Nagao & Matsuyama [1], an estimation of the straightness of a region's boundary was implemented as follows. First, the boundary of the region was traced by means of the standard boundary tracing algorithm attributed to Moore [87], [88]. This results in a sequence of pixels describing a closed walk along the region's outer boundary, denoted as $p_i$, with $i = 1,…, n$, where $p_1 = p_n$ (since the boundary is closed), so that the total number of boundary pixels is $n$ - 1. For each pixel $p_i$ on the boundary, the angle $\Delta i$ between the two lines connecting $p_i$ with $p_{i-s}$ and $p_{i+s}$ was calculated, where values ($i$ + $s$) and ($i - s$) are modulo $n$, i.e., they belong to range $\{0, …, n-1\}$. Variable $s$ is referred to as the step size. A pixel $i$ was counted as "straight" if $|\Delta i| \le \alpha$, for some angle threshold $\alpha$, which is given as a parameter to the method. Let $n_s \in \{0, …, n\text{-}1\}$ denote the number of straight pixels in the boundary, measured using a step size s. Then, the straightness of boundary for step size $s$ is *Straightness$_s$* = $n_s$ / ($n$ - 1) $\in$ [0, 1]. In practice, the step size $s$ acts as an observation scale. To deal with images of different resolution, as well as geometric objects of different size, it is important to choose an appropriate value of the step size $s$. Whereas Nagao & Matsuyama worked with a single step size, the following heuristic criterion was adopted to infer a step size adaptively. First, variable *Straightness$_s$* was computed for all values of $s \in \{1, 2, 4, 8, 16, 32\}$, where dyadic (power of 2) scale values in pixel unit were chosen for selection. Next, the final straightness measure was taken as the one with the maximum value, such that:

$$MltSclStrghtnsOfBndrs = \max_s \{Straightness_s \in [0, 1], \text{ with } s = 1, 2, 4, 8, 16, 32\}. \qquad (4)$$

To summarize, due to its multi-scale implementation, the proposed *MltSclStrghtnsOfBndrs* estimator proved to be less sensitive to segmentation noise than its traditional single-scale counterpart, proposed by Nagao & Matsuyama [1].

*6)  Pixel- and segment-based morphological multiscale characteristic*

Spatial autocorrelation is of fundamental importance for human low-level vision, with special regard to the full primal sketch [3], which is responsible of image texture segmentation (perceptual grouping) [11], [12], [13], [14]. As a consequence, spatial autocorrelation is or should be important in the computer vision and RS common practices [95], [96]. For example, in the RS literature spatial autocorrelation has been proposed as one quality term in a multi-objective $Q^2A$ function for image segmentation [97]. Last but not least, spatial autocorrelation is equivalent to the Tobler's first law of geography (TFLG) [98], considered the foundation of geostatistics and geographical sciences [99], [100], [101]. The TFLG states that, in the geospatial domain, "everything depends on everything else, but closer things more so" [98], although certain phenomena clearly constitute exceptions [102].  In practice, the TFLG implies that a spatial distance decay function exists, such that even though all geospatial observations have an influence on all other observations, after some distance threshold that influence can be neglected. Unfortunately, according to Lembo, "many geographers would say 'I don't understand spatial autocorrelation'; actually, they don't understand the mechanics, they do understand the concept" [91].

In their image segmentation algorithm, Pesaresi and Benediktsson computed a pixel-wise morphological multiscale characteristic, defined as the morphological scale where each pixel's DMP scores its maximum [85]. In our understanding, the morphological multiscale characteristic belongs to the class of local indicators of spatial association (LISA) [86], where spatial association is a synonym of spatial autocorrelation. Unlike existing LISA proposed in the computer vision literature [86], the morphological multiscale characteristic formulated by Pesaresi and Benediktsson is simultaneously pixel-specific and edge sensitive [85]. In practice, it provides a per-pixel estimate of the local size (in pixel unit) of the image-object that pixel belongs to, without requiring spatial units (image-objects) to be detected beforehand by means of an image segmentation algorithm. Any inductive data learning algorithm for image segmentation (or, vice versa, image-contour detection) is inherently ill-posed [11]; hence, it is semi-automatic (where its degree of automation is monotonically decreasing with its number of system's free-parameters, to be user-defined based on heuristics) and site-specific (data-dependent) [56], refer to Section V.A. Since it requires no image segmentation first stage, the multiscale morphological characteristic, adopted as a LISA, is expected to feature $Q^2IO$ values, including computation efficiency and degree of automation, higher than those of traditional LISA, such as Moran's I and Geary's C, reviewed hereafter for comparison purposes. The well-known Moran's I spatial autocorrelation index is either global (image-wide) or a LISA, which means a spatial autocorrelation value specific for each so-called "spatial unit" belonging to an image partition (segmentation) defined beforehand [86].

$$Moran's\ I = \frac{N}{\sum\limits_{i=1}^{N}\sum\limits_{j=1}^{N} w_{i,j}} \frac{\sum\limits_{i=1}^{N}\sum\limits_{j=1}^{N} w_{i,j}\left(X_i - \overline{X}\right)\left(X_j - \overline{X}\right)}{\sum\limits_{i=1}^{N}\left(X_i - \overline{X}\right)^2}$$

$$= \frac{N}{2W} \frac{\sum\limits_{i=1}^{N}\sum\limits_{j=1}^{N} w_{i,j}\left(X_i - \overline{X}\right)\left(X_j - \overline{X}\right)}{\sum\limits_{i=1}^{N}\left(X_i - \overline{X}\right)^2} \in \left[-1,1\right],$$

(5)

$$Local\ Moran's\ I_i = \frac{1}{\sum\limits_{j=1}^{N} w_{i,j}} \frac{\sum\limits_{j=1}^{N} w_{i,j}\left(X_i - \overline{X}\right)\left(X_j - \overline{X}\right)}{\sum\limits_{i=1}^{N}\left(X_i - \overline{X}\right)^2},\ i = 1,...,N,$$

(6)

where $N$ is the number of spatial units indexed by $i$ and $j$, X is the variable of interest, $\overline{X}$ is the mean of X, $w_{i,j}$ is a matrix of spatial weights monotonically non-increasing with distance between spatial units $i$ and $j$. For example, $w_{i,j} = 1$ if spatial units $i$ and $j$ are adjacent and $w_{i,j} = 0$ otherwise. Coefficient $W$ is the sum of all spatial weights $w_{i,j}$. Moran's I ranges from -1 (perfect inverse autocorrelation) to +1 (perfect positive autocorrelation). A zero value indicates a random spatial pattern. For example, in a black-and-white chessboard, where each square, either black or white, is a spatial unit and where the matrix of spatial weights consists of 1s in a 4-adjacency neighborhood and 0s otherwise, then the Moran's I value is equal to -1. According to Equation (6), there is one LISA estimate for each $i$-th spatial unit in the dataset. The sum of LISA values for all spatial units in an image region of interest is proportional to a corresponding global indicator of spatial association for that dataset. By "decomposing" a global autocorrelation result into its local parts, LISA values are very useful to uncover hidden, local patterns in data that the global statistics average over. For example, LISA values can detect when a significant global autocorrelation statistic at a given spatial lag may hide large spatial patches of no autocorrelation, or when an insignificant global autocorrelation statistic may hide patches of autocorrelation.

Moran's I is inversely related to another well-known global spatial autocorrelation index, Geary's C.

$$Geary's\ C = \frac{N}{2\sum\limits_{i=1}^{N}\sum\limits_{j=1}^{N} w_{i,j}} \frac{\sum\limits_{i=1}^{N}\sum\limits_{j=1}^{N} w_{i,j}\left(X_i - X_j\right)^2}{\sum\limits_{i=1}^{N}\left(X_i - \overline{X}\right)^2}$$

$$= \frac{N}{2W} \frac{\sum\limits_{i=1}^{N}\sum\limits_{j=1}^{N} w_{i,j}\left(X_i - X_j\right)^2}{\sum\limits_{i=1}^{N}\left(X_i - \overline{X}\right)^2} \in [0,2),$$

(7)

$$Local\ Geary's\ C_i = \frac{1}{2\sum\limits_{j=1}^{N} w_{i,j}} \frac{\sum\limits_{j=1}^{N} w_{i,j}\left(X_i - X_j\right)^2}{\sum\limits_{i=1}^{N}\left(X_i - \overline{X}\right)^2},\ i = 1,...,N.$$

(8)

The value of Geary's C lies between 0 and values $\geq 2$. Value 1 means no spatial autocorrelation. Values lower than 1 demonstrate increasing positive spatial autocorrelation, whilst values higher than 1 illustrate increasing negative spatial autocorrelation. A simple reversed Geary's C formulation, such that

$$\text{Reversed Geary's C} = 1 - min\{\text{Geary's C}, 2\}, \tag{9}$$

ranges from -1 (indicating inverse correlation) to +1 (perfect autocorrelation), like the Moran's I index.

Our understanding of the morphological multiscale characteristic as a viable alternative to existing LISA estimates, such as Equation (6) and Equation (8), agrees with some statements reported by Pesaresi and Benediktsson in their original paper [85]. An illustrative example of the behavior of the morphological multiscale characteristic is presented in Fig. 4 where, for each geometric object, the average of the characteristic scale over its pixels is adopted as an attribute describing the within-segment average of per-pixel LISA. The definition of a per-pixel DMP requires definitions of opening and closing by reconstruction for a grayscale image $I$. An *opening by reconstruction* is defined as $\gamma^*_\lambda I = \text{Rec}(\varepsilon_\lambda I, I)$, where $\varepsilon_\lambda I$ denotes an erosion of $I$ with a structuring element (SE) of size $\lambda$ and $\text{Rec}(\varepsilon_\lambda I, I)$ denotes the reconstruction by dilation of $I$ from $\varepsilon_\lambda I$. For a detailed formal definition of the erosion and reconstruction operations, refer to [85], [103]. The traditional morphological opening of an image by a SE of size $\lambda$ is used to filter out bright structures that are smaller than $\lambda$. The opening by reconstruction operator also filters out bright structures smaller than $\lambda$, but without affecting the fine-scale details of larger structures, since these are recovered in the reconstruction step. Analogously, a *closing by reconstruction* is defined as $\phi^*_\lambda I = \overline{\text{Rec}}(\delta_\lambda I, I)$, where $\delta_\lambda I$ denotes a dilation of $I$ with an SE of size $\lambda$ and $\overline{\text{Rec}}(\delta_\lambda I, I)$ denotes the reconstruction by erosion of $I$ from $\delta_\lambda I$. Analogously to the opening by reconstruction, a closing by reconstruction filters out dark structures that are smaller than $\lambda$. Starting from these definitions, the opening profile of an image $I$ is composed of a series of openings by reconstruction with a dyadic sequence of SE sizes, $\lambda_i$ for $i = 0, \ldots, n$, such that $\lambda_0 = 0$, $\lambda_1 = 1$, and $\lambda_i = 2^{i-1} + 1$ for $i = 2, \ldots, n$. As an example of the sequence of sizes, if $n = 4$ then the resulting sequence is $\lambda_i = 0, 1, 3, 5, 9$. Differently from Pesaresi and Benediktsson [85], we chose dyadic SE sizes $\lambda_i$, considered to be more practical (and biologically more plausible [104]), so that the resulting profile may be computed in a reasonable amount of time while being able to handle image structures of varying sizes. Given this sequence of spatial scales, the *opening profile* at pixel $x$ is defined as the vector

$$\Pi\gamma(x) = \{\Pi\gamma_{\lambda_i} : \Pi\gamma_{\lambda_i} = \gamma^*_{\lambda_i}(x), i = 0, \ldots, n\}. \tag{10}$$

Correspondingly, the *closing profile* at pixel $x$ is

$$\Pi\phi(x) = \{\Pi\phi_{\lambda_i} : \Pi\phi_{\lambda_i} = \phi^*_{\lambda_i}(x), i = 0, \ldots, n\}. \tag{11}$$

The per-pixel DMP records the rate of change in the opening and closing profiles. For each pixel, its DMP provides an estimate of the importance of structures of size $\lambda$ to which the pixel might belong. The *derivative of the opening profile* $\Delta\gamma(x)$ and the *derivative of the closing profile* $\Delta\phi(x)$ are defined respectively as

$$\Delta\gamma(x) = \left\{\Delta\gamma_{\lambda_i} : \Delta\gamma_{\lambda_i} = \frac{|\Pi\gamma_{\lambda_i} - \Pi\gamma_{\lambda_{i-1}}|}{(\lambda_i - \lambda_{i-1})}, i = 1, \ldots, n\right\}, \tag{12}$$

$$\Delta\phi(x) = \left\{\Delta\phi_{\lambda_i} : \Delta\phi_{\lambda_i} = \frac{|\Pi\phi_{\lambda_i} - \Pi\phi_{\lambda_{i-1}}|}{(\lambda_i - \lambda_{i-1})}, i = 1, \ldots, n\right\}. \tag{13}$$

The complete DMP is obtained by concatenating the two DMP components expressed above. In order to define the *morphological multiscale characteristic* [85], the *multiscale opening characteristic* and the *multiscale closing characteristic* must be first defined. The *multiscale opening characteristic* $\Phi\gamma(x)$ of an image $I$ at pixel $x$ is the SE size at which the opening DMP takes on the largest value,

$$\Phi\gamma(x) = \{\lambda : \Delta\gamma_\lambda(x) = \vee \Delta\gamma(x)\}, \tag{14}$$

where $\vee$ denotes the supremum. Analogously, the *multiscale closing characteristic* is the SE size for which the closing DMP has its maximum value,

$$\Phi\phi(x) = \{\lambda : \Delta\phi_\lambda(x) = \vee \Delta\phi(x)\}. \tag{15}$$

The morphological multiscale characteristic $\Phi(x)$ is chosen as the scale at which the DMP is maximum, whether taken from the opening or closing profile. It can be defined as

$$\Phi(x) = \begin{cases} \Phi\gamma(x): \vee \Delta\gamma(x) > \vee \Delta\phi(x) \\ \Phi\phi(x): \vee \Delta\gamma(x) < \vee \Delta\phi(x) \\ 0 : \vee \Delta\gamma(x) = \vee \Delta\phi(x) \end{cases}. \tag{16}$$

In the rest of this paper, variable $\Phi(x)$ is identified as *DMPmltSclChrctrstc*. An illustrative example of applying this definition to all pixels of an image is presented in Fig. 4. Hence, *DMPmltSclChrctrstc*, which is computed pixel-wise, can be averaged per 2D shape.

 A practical challenge in using the pixel-based morphological multiscale characteristic value is that computing openings and closings by reconstruction for large images can be prohibitively slow. In order to speed up this process, two techniques were selected from the existing literature: decomposable filters for dilation and erosion [105], and the downhill filter for reconstruction [106]. Hereafter, the sole case of opening by reconstruction is discussed, because its dual problem is closing by reconstruction for which the same observations hold. The initial erosion operation $\varepsilon_\lambda I$, that occurs before performing the reconstruction by dilation $\mathrm{Rec}(\varepsilon_\lambda I, I)$, can be very fast depending on the structuring element that is used. Instead of using a disk, which seems like the natural choice, it is much faster to approximate the result that would be obtained from a disk by using a regular polygon [107]. In particular, erosion with a square structuring element is extremely fast, since it can be decomposed into single horizontal and vertical operations [105]. The purpose of the initial erosion operation is to eliminate bright structures that are smaller than the structuring element size. The reconstruction step that follows will guarantee that finer scale details of the remaining structures are reconstructed. In practice, the final results obtained from opening by reconstruction were found to be similar whether using a square or a disk for the initial erosion operation. To recapitulate, in the present study, the use of a square structuring element was preferred due to its superior implementation simplicity and lower computation time.

 Due to computation time considerations, special attention must be given to the reconstruction by dilation operation, $\mathrm{Rec}(\varepsilon_\lambda I, I)$. Vincent defined the notion of morphological reconstruction for grayscale images [108] as well as fast algorithms to speed up execution times considerably [103], [108]. Later, Robinson and Whelan presented the downhill filter for image reconstruction [106], with a precondition to be satisfied by input images in order to guarantee correctness. In the weak form, the precondition requires that, in the case of reconstruction by dilation, the marker image be everywhere less than or equal to the mask image. Fortunately, in the reconstruction by dilation $\mathrm{Rec}(\varepsilon_\lambda I, I)$, this condition always holds true, since $\varepsilon_\lambda I \leq I$. The downhill filter computes the reconstruction in a single pass, guaranteeing a fast and consistent execution time. In experimental comparisons with the algorithms proposed by Vincent [103] over a range of input images, the downhill filter showed a consistent and oftentimes large speedup [106]. Hence, we adopted the downhill filter to compute the required reconstructions.

 To summarize, the implemented morphological *DMPmltSclChrctrstc* function is proposed as a computationally efficient pixel-wise LISA sensitive to the presence of local edges. Unlike existing LISA estimates, such as the popular local Moran's I and Geary's C formulations, see Equation (6) and Equation (8), *DMPmltSclChrctrstc* requires no *a priori* partition (segmentation) of an image into spatial units (image-objects), which improves $Q^2IO$ values of the LISA estimator, including degree of automation and robustness to changes in the input image.

<p align="center">(Fig. 7 about here)</p>

*7) Area-based elongatedness sensitive to within-segment holes*

A new measure of elongatedness is presented to overcome operational limitations of the three formulations proposed by Nagao & Matsuyama, each at an increasing level of sophistication [1]. Their most sophisticated solution is summarized below, to be compared with our new measure. To define a measure of elongatedness, Nagao & Matsuyama proposed to estimate the longest path along a region, together with the region's average width along that path [1]. First, holes of the region are filled in, if any, after which the region's skeleton is computed via thinning. Next, the longest path along the skeleton is detected. Additionally, for each point along the longest path, the local width of the region is computed and the average of these widths is taken. The elongatedness measure defined by Nagao & Matsuyama (NM) is

$$Elngtdnss_{NM} = L_{NM} / W_{NM}, \qquad (17)$$

where $L_{NM}$ is the length of the longest path and $W_{NM}$ is the average width along that longest path. A practical limitation of the *Elngtdnss$_{NM}$* operator is that filling in the holes of a region before computing its elongatedness may not always be appropriate. Fig. 7 shows examples of segments of roads and rivers, extracted from a satellite image, where perception of elongatedness by a human expert scores "high", whereas index *Elngtdnss$_{NM}$* does not, due to preliminary within-segment hole filling required by Nagao & Matsuyama's procedure. This limitation justifies the introduction of an improved measure of elongatedness, different from Nagao & Matsuyama's, discussed hereafter. Within-segment holes, if any, are not filled in beforehand. After computing the region's complete skeleton (without removing holes), our novel measure of elongatedness is defined as

$$ElngtdnssAndNoHole = L / W \geq 1, \qquad (18)$$

where L is a measure of the length of the complete skeleton (without removing holes), i.e., L is the total number of skeleton pixels, while W is the region's average width across each pixel along the skeleton. The novel measure of *ElngtdnssAndNoHole* captures a different notion of elongatedness from index *Elngtdnss$_{NM}$*. First, in the former, the length measure L is estimated as the number of pixels in the whole skeleton, as opposed to just its longest path. This way, L takes into consideration every part of a region. This difference is illustrated in Fig. 8. Second, by not filling in possible holes, variable *ElngtdnssAndNoHole* treats holes as intrinsic characteristics of image-objects, as opposed to ignoring them. Fig. 9 shows how the two measures of elongatedness behave differently on a toy region with a hole. Another important difference between variables *ElngtdnssAndNoHole* and *Elngtdnss$_{NM}$* is the algorithm used to compute the region's skeleton. Rather than a traditional thinning-based skeletonization algorithm, like the one adopted by Nagao & Matsuyama [1], we adopted an algorithm for the estimation of the region's filtered Euclidean skeleton. Many thinning-based skeletonization algorithms were proposed and analyzed over the years [109]. In general, direct thinning skeletonization procedures, though formulated with certain desirable properties in mind, present results that are not always easily predictable, i.e., results not in line with expectations. In recent years, new developments have been achieved in techniques for fast computation of Euclidean skeletons. These new techniques ensure a correct topology of the detected skeleton and allow for filtering out eventual noise [110]. An important advantage of using Euclidean skeletons is that they are better defined: in principle, a Euclidean skeleton can be defined simply as the set of points centered in the shape with respect to the Euclidean distance[5]. In addition, in common practice, Euclidean skeletonization algorithms compute faster than traditional region thinning algorithms. We obtained a filtered Euclidean skeleton by selecting Couprie *et al.*'s method [110]. A problem with the resulting skeleton is that it may be 4-adjacency connected. For our purposes, to obtain more consistent measurements of the total length L of the skeleton, skeleton pixels should be 8-adjacency connected. To obtain the final 8-adjacency connected skeleton we applied a cycle of the Cychosz's [111] fast implementation of Rosenfeld's parallel thinning algorithm [112], [113]. As shown in proposition 6 of the paper by Rosenfeld [112], this guarantees the final result is 8-adjacency connected. Finally, to compute the average width W of the skeleton, an original implementation was adopted to estimate the region's width across each pixel along the skeleton. As an intermediate product necessary to compute the Euclidean skeleton, Couprie *et al.*'s method computes the Euclidean distance transform of the region by using the linear-time method proposed by Meijster *et al.* [114], [115]. To estimate W in near real-time, the required per-pixel distances were read from the pre-computed values of the Euclidean distance transform.

In summary, in our experiments the proposed *ElngtdnssAndNoHole* index formulation and implementation improved robustness to noisy data and reduced computation time in comparison with its existing counterparts discussed by Nagao & Matsuyama [1].

<div align="center">(Fig. 8 about here)</div>
<div align="center">(Fig. 9 about here)</div>

*8)  Area-based simple connectivity as a measure of the presence of within-segment holes*

Within-segment holes can be considered a shape property important in the identification of different classes of real-world objects depicted in images. From a perceptual standpoint, Bertamini showed that it is important to consider holes as constituent features of their enclosing objects [116]. Besides being a perceptually relevant geometric feature *per se*, holes affect area-based geometric attributes estimated in the spatial domain. On the one hand, contour-based shape descriptors are traditionally considered more sensitive to noise and variations than area-based geometric descriptors, assuming the latter are employed once within-segment holes, if any, have been filled in. On the other hand, unlike contour-based geometric attributes, area-based shape descriptors are affected by holes. If an object-specific simple connectivity value scores low, then "simple" (intuitive to use) area-based geometric indexes in the spatial domain are biased by the presence of holes and should be dealt with special care by an OBIA system. For example, inner holes in an image-objects increase geometric feature *ElngtdnssAndNoHole* and decrease geometric features *Convexity* and *Roundness*. In common practice, within-segment holes may be due to segmentation errors (e.g., undersegmentation phenomena) occurring in the inherently ill-posed image segmentation first stage (raw primal sketch [3]). In this case, segment holes can be considered an indication of segmentation noise.

To quantify the extent to which a segment is simply connected, i.e., the degree to which the segment is free of holes, an original normalized simple connectivity measure was designed and implemented as an indicator of the degree of confidence of area-based geometric indexes in the spatial domain [117]. Unfortunately, there is a limited amount of previous works on measures that assess the presence of holes. The simplest measure is the absolute number of holes of a region or, otherwise, its Euler number [7], [88]. Although the number of within-segment holes is informative, it gives no clue about the holes' individual size and position and their overall extent and spatial distribution. Soffer & Samet [118] and Wentz [119] defined geometric measures based on the area of holes relative to that of the region. A disadvantage of area-based measures of simple connectivity is their independence from the holes' spatial distribution. This justifies our presentation of a

---

[5] In spite of the simplicity of its mathematical definition, in practice, computing Euclidean skeletons is quite complicated due to the discrete nature of images.

novel simple connectivity measure estimated as a fuzzy-AND (minimum) combination of two quantitative terms, one contour- and one area-based. The first term, called *SmplCnctvty4Adjcncy*, is contour-based. It quantifies the presence of holes by relating the length of the boundaries of the holes to the total length of all boundaries (equal to outer boundary + inner boundaries, if any) of the region. It is defined as:

*SmplCnctvty4Adjcncy* = 4-adjacency cross-aura measure of the external boundary / 4-adjacency cross-aura measure of the total boundary, (19) whose numerator, the "4-adjacency cross-aura measure of the external boundary" does not take into account inner boundaries of holes, if any. On the contrary, the denominator "4-adjacency cross-aura measure of the total boundary" does take into account contributions from holes. Hence, *SmplCnctvty4Adjcncy* belongs to range [0, 1]. A detailed explanation on how these boundary lengths are computed is presented in [117]. The disadvantage of the *SmplCnctvty4Adjcncy* term is its lack of sensitivity to the presence of one or few holes with a large area, but small boundary lengths. In this situation, where a desirable simple connectivity variable is expected to score low, *SmplCnctvty4Adjcncy* scores some intermediate value, which is perceptually counter-intuitive, see Fig. 10. This problem motivates the fuzzy-AND (minimum) combination of the *SmplCnctvty4Adjcncy* measure with an area-based geometric measure, called *FilledAreaRatio*, defined as

*FilledAreaRatio* = Area / FilledArea, (20)

where FilledArea is the area of the region with its holes filled in. Hence, *FilledAreaRatio* belongs to range [0, 1]. The disadvantage of this second term of the simple connectivity variable is its low sensitivity to the presence of multiple small holes, i.e., holes featuring overall a large total boundary but a small total area, see Fig. 10. The proposed final measure of simple connectivity is the conservative (fuzzy-AND) combination of the two previously defined fuzzy membership values,

*CombndSmplCnctvty* = Fuzzy-AND{ *SmplCnctvty4Adjcncy*, *FilledAreaRatio*} = Min{ *SmplCnctvty4Adjcncy*, *FilledAreaRatio*}, (21)

where *CombndSmplCnctvty* belongs to range [0, 1]. When there is no hole in the region, *CombndSmplCnctvty* scores 1. Fig. 10 shows how this ultimate simple connectivity formulation performs in better agreement with our perception of holes in a region.

(Fig. 10 about here)

*B.  Pairwise Feature Test of Statistical Independence*

Seven geometric measures, featuring twenty-one pairwise combinations (refer to Section VI.A) were analyzed for statistical independence by means of the Pearson's chi-square test for independence, in compliance with statistical test constraints listed in Section V.B.1. First, the two test populations of real-world geometric objects, described in Section IV, were subject to simple random sampling by a decimation factor (1/10). Next, the two sample sets were added to the third synthetic dataset to form a single test set, consisting of 745 samples. According to Equation (1), an empirical choice to transform a quantitative variable into a qualitative variable, eligible for being input to the Pearson's chi-square test for independence, is to choose k = number of equiprobable buckets = $2 * N^{(2/5)}$, where N = finite size of the sample dataset = 745, therefore k = 28 [82]. Hence, the cumulative distribution of each sample variable was estimated to detect twenty-eight equiprobable buckets per variable. Twenty-one two-way contingency tables were computed between pairs of the seven quantized geometric features, together with their expected frequency counts. The expected frequency count for each cell of the contingency tables was at least 5, to comply with the statistical analysis requirements, refer to Section V.B.1. Finally, the chi-square test statistic was computed with the Windows Excel CHITEST() function, whose output is the P-value (refer to Section V). In addition to the Pearson's chi-square test for independence, the normalized Pearson's chi-square index [50], also known as Cramer's V index (CVI) [52], was estimated, in agreement with [120]. It is defined as follows.

CVI = Pearson's chi-square index / Maximum of the Pearson's chi-square index = Pearson's chi-square index / [N * (min(row, column in the contingency table) – 1)], CVI    (22) $\in$ [0, 1], where N is the number of samples [120]. Results are shown in Table 1 and Table 2 respectively. About the P-value and the CVI results, the following considerations hold.

<div align="center">(Table 1 about here)</div>
<div align="center">(Table 2 about here)</div>

(i)   The P-value is the probability that a chi-square statistic having (row – 1) * (column – 1) degrees of freedom is equal or superior to the chi-square statistic value computed in the sample test for independence. When the P-value is less than the significance level $\alpha$, fixed equal to 0.05, then the null hypothesis of independence, H0, cannot be accepted at a level of confidence equal to 0.95 (refer to Section V). The P-value features an inductive inference value, i.e., it is suitable for use in an inductive (bottom-up, data-driven) inference framework, to make predictions upon the entire population based on a population sample. In common practice, chi-squared values tend to increase (showing increasing dependence) with the number of contingency cells. This may be due to any of the accidental or systematic errors listed in [50].

(ii)   The CVI may be considered as the association (dependence) between two variables as a percentage of their maximum possible association. CVI varies from 0, corresponding to no association (independence) between the two variables, to 1, meaning complete association [52], [120], [121]. As such, the CVI holds a somehow "absolute" (data-independent) value in the normalized range of change [0, 1]. In practice, to gain normalization of its domain of change the CVI sacrifices its sensitivity to changes in the input dataset [52]. It is known that a discrepancy may arise in statistical sampling when the chi-square P-value turns out to be very low, such that the null hypothesis of independence between the two categorical input variables is dismissed based on evidence collected from sample data. In this situation, the CVI can border zero as an expression of an independence condition, in disagreement with the P-value [120]. On the other hand, in common practice, the greater the difference between rows and columns in the contingency table, which means the smaller the denominator of the CVI formulation, the more likely CVI will increase tending to 1, i.e., it can show increasing dependence although no chi-square statistical dependence at a given significance level is detected. In general, CVI may not be completely accurate for comparing the degree of association in different contingency tables [52]. As a consequence of these known drawbacks, the CVI was considered as yet-another source of weak statistical evidence, but little useful in practice. In fact, no known cut-off value was applied to CVI in Table 2 to mark pairs of random variables as independent.

For the sake of completeness, we mention that, in line with theoretical expectations, the two shape properties omitted from Table 1 and Table 2, namely, area and orientation, proved to be statistically independent from any other proposed variable.

*C. Monotonically Increasing or Decreasing Relationship between Pairs of Planar Geometric Features at the Population and Individual Levels of Analysis*

At the level of analysis of an entire population of 2D shape instances, Table 3 presents the SRCC values estimated in the twenty-one pairwise comparisons of seven ranked variables generated from the planar shape features discussed in Section VI.B.

For each pair of geometric features considered in Table 3, with special regard to cells depicted in dark gray which deserve further investigation to avoid inter-feature causality, one or more 3-tuples of planar objects, R, E1 and E2, were found in the test dataset, capable of fulfilling the third sufficient (and necessary?) condition for feature pair non-causality, refer to Section V.B.3.

<div align="center">(Table 3 about here)</div>

<div align="center">VII. DISCUSSION</div>

*A. Qualitative Analysis of 2D Shape Indexes*

A graphical user interface (GUI) was specifically developed to allow an expert photointerpreter to assess qualitatively whether geometric indexes, estimated from hundreds of different image-objects, comply with theoretical and perceptual expectations. Screenshots of the GUI employed in real-world vision problems, ranging from satellite Earth observation (EO) image understanding to leaf image classification, are shown in Fig. 11 to Fig. 13. These examples illustrate how a discrete and finite dictionary of quantitative geometric terms, provided with an intuitive physical meaning, can be combined together, in addition to per-object photometric features and inter-object spatial relationships, to develop an application-specific physical model-based decision tree, eligible for use in an OBIA system capable of mimicking human reasoning.

<div align="center">(Fig. 11 about here)</div>
<div align="center">(Fig. 12 about here)</div>
<div align="center">(Fig. 13 about here)</div>

Fig. 11 shows an EO image subset where geometric objects, automatically detected by the SIAM expert system in the 2 m resolution spaceborne MS test image (refer to Section IV), are replaced by values of their geometric attributes, specifically *ElngtdnssAndNoHole*, *CombndSmplCnctvty* and *MltSclStrghtnsOfBndrs*. Based on theoretical considerations (refer to Section VI.A.8), if *CombndSmplCnctvty* scores low then area-based geometric indexes in the spatial domain, including *ElngtdnssAndNoHole*, *CnvxtyAndNoHole* and *RndnssAndNoHole*, should be considered biased. This is correctly shown in Fig. 11, where segments containing holes and scoring "low" in *CombndSmplCnctvty*, refer to Fig. 11(c), present somewhat "high" values of *ElngtdnssAndNoHole* in Fig. 11(b), even though they may look relatively compact. The *MltSclStrghtnsOfBndrs* estimate, shown in Fig. 11(d), appears suitable for capturing manmade surface structures, which usually present straight boundaries irrespective of their shape, either simple or complex.

Fig. 12 presents a screenshot of the GUI showing some of the geometric objects detected by the SIAM expert pre-classifier in the spaceborne image depicted in part in Fig. 11(a). Segments numbered 1 through 6 are buildings or parts of buildings, while segments 7 through 9 are pieces of roads. To distinguish buildings from roads in a GEOBIA framework, some general decision rules can be inferred. The most discriminative geometric attribute appears to be *ElngtdnssAndNoHole*. Though buildings can be somewhat elongated, roads usually present very high values of *ElngtdnssAndNoHole*. Additionally, roads are typically non-compact, resulting in lower values of *RndnssAndNoHole*. Regarding appearance, bright segments (e.g., provided with high values of a photometric attribute called "mean panchromatic intensity", complementary not alternative to geometric attributes) are usually indicative of buildings. Though buildings can appear as either bright or dark structures, asphalt roads are always dark, so that bright segments are highly indicative of non-road structures. Noticeably, not all rectangular segments are buildings, as exemplified by the mostly rectangular road segment 9. Some more specific decision rules employing different geometric attributes can be inferred from Fig. 12. Segments 4 and 5 represent buildings whose change in brightness generate holes in their surface area, lowering their simple connectivity values. Nonetheless, both segments have high values of indexes fuzzy rectangularity and straightness of boundaries, which are both contour-based. Segment 6 is particularly interesting, since it represents a building, yet it has an overall concave shape. This causes low values in *CnvxtyAndNoHole* and *RndnssAndNoHole*, though the segment scores high in *MltSclStrghtnsOfBndrs*. Segment 7 depicts a piece of a road network, which results in a very low value of *CnvxtyAndNoHole*. This is not the case of segment 8, which belongs to a single straight piece of road. In common, both segments share a very high value of index *ElngtdnssAndNoHole*, distinguishing them from other geometric objects. Finally, segment 9, though originating from a piece of road, is very hard to be distinguished from a building in general. It is clear that segment 9 depicts a piece of road only if its spatial neighborhood (not shown in Fig. 12) is observed in the image domain: this spatial context consists of road-like segments that possess the same overall orientation and cast no shadow, whereas neighboring buildings tend to cast shadows.

Fig. 13 presents a GUI screenshot of leaves, each from a different tree species. For example, in segment 1, the cluster of pine leaves stands out for having a very small area, high *ElngtdnssAndNoHole*, and low *RndnssAndNoHole*. Segments 2 through 4, which depict compound leaves, share high values of *ElngtdnssAndNoHole*. In particular, segment 2, featuring very thin leaflets, presents the highest *ElngtdnssAndNoHole* value. Compound leaves also tend to present holes formed from the overlap of separate individual leaflets, as exemplified by segments 2 and 3. This results in a decrease of their *CombndSmplCnctvty* index. Segments 6 and 7 represent two different species of oak trees, though the marked protrusions in segment 7 give rise to lower values of *Convexity*, *RndnssAndNoHole* and *MltSclStrghtnsOfBndrs*. The sycamore leaf in segment 9 can be distinguished from the maple in segment 8 mainly by its low *MltSclStrghtnsOfBndrs value*. Finally, segment number 5 is a simple leaf with a smooth boundary, shown here to provide yet another reference set of attribute values, eligible for use in quantitative shape matching and retrieval.

*B. Qualitative Interpretation of Quantitative Statistical (In)dependence Results*

According to Section I, care must be taken not to over-interpret or over-rely on the findings of GOF tests, like the Pearson's chi-square test for independence. GOF tests have notoriously low power and are generally best for rejecting poor distribution fits rather than for identifying good fits [60]. That said, Table 1 shows twenty-one inter-feature P-values computed with the Microsoft Excel CHITEST() function, adopted to finalize the Pearson's chi-square test for independence. If a P-value = Probability(chi-square value > test chi-square value) is less than the selected level of significance $\alpha = 0.05$, then the null hypothesis H0 is rejected at a 95% level of confidence (refer to Section V), i.e., the two random variables are statistically dependent based on inductive inference. If two variables are statistically dependent, then knowing the value of one variable *can* help inferring the value of the second variable. For example, if contour-based *FuzzyRuleBsdRctnglrty*, independent of holes, is "high", then area-based *RndnssAndNoHole* and *ElngtdnssAndNoHole* are expected to be low; if *RndnssAndNoHole* and/or *Elongatedness* are high, then *FuzzyRuleBsdRctnglrty* is expected to be low. As another example, area-based *ElngtdnssAndNoHole,* increasing with holes, and area-based *CnvxtyAndNoHole*, decreasing with holes, are expected to be inversely related irrespective of holes, because an elongated region is likely to be bent, which causes convexity to score low. Based on these preliminary theoretical considerations, all statistical occurrences of (in)dependence highlighted by Table 1 appear intuitive to explain. In Table 1, two-of-seven features, the *DMPmltSclChrctrstc* and the *MltSclStrghtnsOfBndrs*, are statistically independent from any other geometric index. According to these authors' opinion, in Table 1, only one pairwise result of independence was somehow counterintuitive: area-based *CnvxtyAndNoHole*, decreasing with holes, and *CombndSmplCnctvty*, also decreasing with holes, resulted to be statistically independent, whereas we expected this feature pair to be statistically dependent.

It is known that a discrepancy between the chi-square P-value and the CVI may arise in statistical sampling when the chi-square P-value, which has a "relative" data-dependent inference value, turns out to be very low, meaning that the two categorical input variables are dependent. In this situation, the CVI, which belongs to an "absolute" domain of change, can border zero as an expression of a condition of independence, in disagreement with the P-value [120], refer to Section VI.B. This is exactly what happens in Table 2, where all pairwise feature tests show independence in "absolute" terms, irrespective of the "relative" evidence of dependence shown in Table 1. Although the result shown by Table 2 is highly desirable (all variable pairs are independent), it cannot be considered "ultimate", since the findings of GOF tests must always be scrutinized with care by an HDM, based on visual inspection of the probability plots and other comparisons [60]. For example, according to [52], the use of CVI is not recommended because, in the normalization of its domain of change, it sacrifices sensitivity to changes in the input dataset.

## C.  Qualitative Interpretation of Quantitative SRCC results

Table 3 shows SRCC values perfectly in line with theoretical expectations. There are five-of-twenty-one SRCC instances to be considered either "average" or "high", i.e., superior to 0.4 (refer to Section VI.B). All these cases regard area-based geometric features affected by the presence (or absence) of holes, estimated via parameter *CombndSmplCnctvty*, specifically, *ElngtdnssAndNoHole* (monotonically decreasing with *CombndSmplCnctvty*) and *RndnssAndNoHole* (monotonically increasing with *CombndSmplCnctvty*), whereas *RndnssAndNoHole* is monotonically decreasing with *ElngtdnssAndNoHole* and monotonically increasing with *CnvxtyAndNoHole*, which implies that the *CnvxtyAndNoHole* index too is monotonically decreasing with *ElngtdnssAndNoHole*.

Five-of-seven variable pairs considered statistically dependent in Table 1 feature an "average" or "high" SRCC value in Table 3, but only two-of-seven variable pairs score "high" in the SRCC, with three geometric features involved: *ElngtdnssAndNoHole*, *RndnssAndNoHole* and *CnvxtyAndNoHole*. Before the third and last level of the hierarchical test for pairwise variable causality proposed in Section VI, these three variables are eligible for being considered the same dependent variable. If a population-wide distribution is collected (at large spatial extent) from local estimates, then local non-stationary statistics may not survive the averaging process. To account for this consideration, a "local" scrutiny by an HDM of the pairwise feature pair at risk is recommended at the third and last level of the proposed analysis for causality. Actually, it is easy to provide evidence, e.g., based on synthetic examples of individual 2D objects, where the three intuitive geometric variables *ElngtdnssAndNoHole*, *RndnssAndNoHole* and *CnvxtyAndNoHole* show their "local" absence of a monotonically increasing or decreasing pairwise feature relationship. More in general, investigations at a local spatial scale revealed that none of the feature pairs in the selected set of seven geometric features is either monotonically increasing or decreasing, as recommended by Section VI.C. This local-scale conclusion does not contradict large-scale SRCC results shown in Table 3, but provides insights on the behavior of planar geometric features at multiple spatial scales of analysis, from local scale, at the level of individuals, to global spatial extent, at the level of population of individuals.

Since implementation of the three levels of sufficient but not necessary criteria, required by Section VI to investigate pairwise feature non-causality, brought no evidence of causal relationship, the selected set of seven 2D shape attributes can be considered minimally dependent (mD), until proved otherwise.

## VIII. Conclusions

The present research and technological development (RTD) software project moves from the seminal works by Nagao & Matsuyama [1], [2] and Shackelford & Davis [68], [69], published in the geographic object-based image analysis (GEOBIA) literature, to design, implement and validate, for quantitative quality assurance ($Q^2A$) purposes, an original general-purpose dictionary of seven off-the-shelf 2D shape descriptors in the spatial domain, provided with an intuitive physical meaning to be suitable for use in (GE)OBIA systems in operating mode, required to mimic human reasoning. This is an inherently ill-posed multi-objective optimization (cognitive) problem. To become better posed for numerical solution, it adopts the following original combination of constraints. The ensemble of planar geometric indexes is required to be: (I) general purpose, (II) minimally dependent (mD) and (III) maximally informative (MI), in compliance with the Occam's razor principle, familiar to the machine learning community (refer to footnote 2), (IV) intuitive to use in a (GE)OBIA paradigm, expected to mimic human reasoning, and (V) subject to a *Val* policy for $Q^2A$, in compliance with the Quality Assurance Framework for Earth Observation (QA4EO) guidelines (refer to Appendix 3). In addition to the ensemble mDMI optimization criteria, each individual 2D shape descriptor is expected to optimize a set of community-agreed quantitative quality indicators ($Q^2I$) of operativeness ($Q^2IOs$), including: (VI) accuracy, (VI) computational efficiency, (VII) invariance with respect to translations, rotations and scaling transformations, and (VIII) robustness to the presence of noise in the data. Also the operator-specific maximization of $Q^2IOs$ is subject to a *Val* policy for $Q^2A$, in compliance with the QA4EO guidelines (refer to Appendix 3).

Unfortunately, but realistically and in line with existing literature, no "objective" (quantitative) assessment of the qualitative degree of informativeness (MI) of the proposed set of application-independent descriptors is claimed in this study, to be rather accomplished by a (subjective) human decision maker (HDM) in his/her own data- and application-specific domain of interest.

Of potential interest to a broad audience of computer vision and RS scientists and practitioners, conclusions of this software project are twofold. First, the proposed general-purpose dictionary of seven off-the-shelf 2D geometric descriptors in the spatial domain (plus area and orientation, refer to Section VI), specifically, *CnvxtyAndNoHole*, *FuzzyRuleBsdRctnglrty*, *RndnssAndNoHole*, *MltSclStrghtnsOfBndrs*, *DMPmltSclChrctrstc*, *ElngtdnssAndNoHole* and *CombndSmplCnctvty*, provided with an intuitive physical meaning and subject to a known *Val* policy for $Q^2A$, can be integrated into software libraries of efficient and reliable computational geometry algorithms, such as CGAL [27], where "simple" 2D shape descriptors are absent. This compact feature set of certified quality is alternative to the large number and variety of 2D shape functions implemented in existing commercial or open source software libraries, such as eCognition's [30], OpenCV [31] (refer to Appendix 1) and ENVI's [32] (refer to Appendix 2), whose multi-objective geometric feature representation and description criteria remain unknown for *Val* purposes. At the level of understanding of system design and knowledge/information representation [3], [15], three of the seven proposed geometric features, specifically, *MltSclStrghtnsOfBndrs*, *DMPmltSclChrctrstc* and *CombndSmplCnctvty* are totally new, i.e., they are absent from the three aforementioned commercial or open source software libraries and from the reference works by Nagao & Matsuyama [1], [2] and Shackelford & Davis [68], [69]. At the level of understanding of algorithms and implementation [3], [15], all the proposed geometric descriptors feature several degrees of novelty, introduced to optimize their $Q^2IO$ values in comparison with alternative solutions.

The second original contribution of the present study of potential interest to a wide scientific audience is the proposed hierarchical *Val* strategy for $Q^2A$ of a set of quantitative random variables whose dependence (causality [48]) must be minimized (mD). At the first sufficient level of investigation for independence, in place of the popular Pearson's cross-correlation coefficient (PCC), which is sensitive to bivariate linear relationships exclusively, a more general Pearson's chi-square test for independence is proved to be feasible and more adequate than PCC to quantitatively investigate the degree of pairwise feature dependence. This statistical improvement comes at the cost of a preliminary transformation of each input quantitative random variable into a qualitative (nominal) variable featuring an adequate number of equiprobable levels of discretization, according to heuristic statistical criteria, see Equation (1). In common practice, the Pearson's chi-square test for independence is implemented by the well-known Windows Excel CHITEST() function, capable of coping with a chi-square distribution whose number of degrees of freedom is very high, equal to (row − 1) * (column − 1), where row = column = number of equiprobable levels of discretization = 28 in our experiments (refer to Section VI.B). At the second sufficient level of investigation, to be applied when the Pearson's chi-square test for independence reveals statistical dependence of a variable pair, the Spearman's rank cross-correlation coefficient (SRCC) value is estimated. It shows whether two ranked random variables are monotonically increasing or decreasing, independent of linear relationships. Finally, at the third level of investigation, to be applied if the first two sufficient tests for independence reveal pairwise variable dependence, a local proof of the absence of monotonically increasing or decreasing pairwise feature relationships is required to account for the typical nonstationarity of planar statistics. Overall, convergence-of-evidence stemming from these three levels of analysis is perfectly in line with theoretical expectations.

Future applications of the implemented set of general-purpose 2D shape descriptors in operating mode will regard specific image domains and (GE)OBIA classification systems, in comparison with alternative software libraries of geometric functions.

APPENDIX 1 - OPENCV LIBRARY

The OpenCV Library [31] supports the estimation of a wide set of quantitative geometric indexes as summary characteristics of contours, rather than area-based polygons, so that two (1D) curves or (2D) polygons, equivalent to, respectively, open or closed contours, can be compared (matched) for (dis)similarities by simply computing their summary characteristics and estimating their difference according to a chosen (dis)similarity criterion. As reported in Section II, area-based descriptors are traditionally considered more robust, i.e., less sensitive to noise or shape deformations, while boundary-based descriptors are considered more sensitive. The OpenCV contour-specific geometric descriptors and matching functions are summarized below.

1. If we are drawing a contour (in vector data format) or are engaged in shape analysis, it is common to approximate a contour representing a polygon with another contour having fewer vertices. This is accomplished with the routine cvApproxPoly(), implemented as the RDP approximation [93].

2. Closely related to the contour/polygon approximation is the process of finding dominant points, with the routine cvFindDominantPoints().

3. Contour length - cvArcLength(); Contour (polygon) area – cvContourArea();

4. Positional attributes as bounding boxes, specifically, cvBoundingRect(): horizontal bounding rectangle; cvMinAreaRect2(): to handle rectangles of any inclination; Enclosing circles and ellipses; Convex hull and convexity defects.

5. One of the simplest ways to compare two contours is to compute contour moments: Moments, Normalized moments (scale-invariant) and the seven Hu invariant moments (scale-invariant, rotation-invariant) – cvGetHuMoments(). Unfortunately, contour moments lack an intuitive meaning.

6. Pairwise geometrical histograms (PGH) as a generalization of a chain code histogram representation of a contour – cvCalcPGH().

7. Several functions provide hierarchical contour trees and accomplish hierarchical matching of contour trees.

APPENDIX 2 – ENVI EX, 5.0

In the ENVI EX commercial software product [32], the feature extraction phase adopts an OBIA approach to classification, as opposed to pixel-based classification. Image-objects can be depicted with a variety of spatial, spectral, and texture attributes, to be selected interactively by the user. If the user chooses to compute spatial attributes, ENVI EX performs an internal raster-to-vector operation and computes spatial attributes from the vectors. In particular, ENVI EX calculates all of its spatial attributes based on a smoothed version of the geometry, not the original geometry, according to the RDP approximation [93]. Performing calculations on a smoothed geometry ensures that shape measurements are less sensitive to object rotation and image noise. The list of spatial descriptors supported by the ENVI EX 5.0 commercial software toolbox is provided in Table 4 [32].

(Table 4 about here)

APPENDIX 3 – QA4EO GUIDELINES

According to the QA4EO recommendations [33], delivered by the intergovernmental Group on Earth Observations (GEO), the visionary goal of a timely, comprehensive and operational transformation of massive amounts of spaceborne/airborne EO images into information products requires the successful implementation of two key principles: Accessibility/Availability and Suitability/Reliability of RS data, processes and outcomes. The QA4EO key principle of Suitability/Reliability relies on mandatory calibration and validation (*Cal/Val*) activities, whose implementation becomes critical to the quantitative (metrological/statistically-based) $Q^2A$ of data, processes and products. *Cal* is the transformation of sensory data into a physical unit of radiometric measure [122], which guarantees harmonization and interoperability of multi-source and/or multitemporal datasets. Val is the process of assessing, by independent means to be community-agreed upon, the "standard" quality of process and outcome. It requires each data processing stage and output product to be assigned with $Q^2Is$, to be community-agreed upon, featuring a degree of uncertainty in measurement at a known degree of statistical significance. Hence, *Val* provides a documented traceability of the propagation of errors through the information processing chain, in comparison with established "community-agreed reference standards" [33].

Quite strikingly, the GEO's *Cal/Val* requirements included in the QA4EO guidelines, although regarded as common knowledge, are neglected or ignored in the RS common practice [15], [16]. About the *Cal* requirement, it is an unquestionable fact that the word "calibration" is absent from a large portion of papers published in the RS literature. For example, when popular spectral indexes, e.g., vegetation indexes, are computed from raw digital numbers not transformed into a radiometric unit of measure, such as top-of-atmosphere reflectance, published conclusions lack any physical meaning, e.g., refer to ([123], p. 3027, Fig. 1). Another unquestionable fact is that, in popular RS image processing commercial software products [30], [32], RS data *Cal* is not a pre-requisite [15], [16]. It means that these popular RS image processing commercial software products are collections of statistical (inductive, bottom-up) rather than physical (deductive, top-down) model-based algorithms. The former are inherently ill-posed [38], [39], semi-automatic and site-specific [56], but do not consider data *Cal* as mandatory, although their robustness to changes in the input dataset may benefit from data harmonization accomplished by a data *Cal* policy. In the RS common practice, EO image understanding systems (EO-IUSs) relying exclusively on inductive learning-from-data algorithms are expected to score low in community-agreed $Q^2$IOs, such as degree of automation, robustness to changes in the input dataset, robustness to changes in input parameters, scalability, transferability and timeliness (from data acquisition to product generation) [15], [16]. About the *Val* requirement, it is a fact that, in the majority of papers published in the RS literature, classification accuracies are not provided with a mandatory degree of uncertainty in measurement [124]; as a consequence, these accuracy values feature no statistical meaning. For example, in the framework of the increasingly popular GEOBIA paradigm [40], a typical EO-IUS implementation, such as that proposed in [125], employs no *Cal* policy; no *Val* strategy is applied to a multi-spectral (MS) image panchromatic sharpening pre-processing and to a low-level inherently ill-posed image segmentation first stage [11]; accuracy estimation of the high-level classification second stage is provided with no degree of uncertainty in measurement, etc. The consequence of this ongoing lack of community-agreed *Val* initiatives, where $Q^2$IOs are estimated in compliance with the QA4EO guidelines, is that the application domain of a great majority of the EO-IUSs published in the RS literature remains unknown to date [15], [16].

In an interdisciplinary framework such as that sketched in Fig. 1, the first QA4EO requirement of *Accessibility/Availability* is related to the well-known Shannon's information theory of data communication [126], called unequivocal ("easy", quantitative) information-as-thing theory by philosophical hermeneutics [4]. The second QA4EO constraint of *Suitability/Reliability* is related to the inherently equivocal ("difficult", qualitative) information theory known as information-as-data-interpretation [4], which is the interdisciplinary focus of attention of epistemology [46], [47] and cognitive science [21], [22], see Fig. 1. Not surprisingly, the first ("easy") QA4EO key principle has recorded significant improvements by the RS community in recent years [127]. Unfortunately, the second ("difficult") QA4EO key principle remains to a large degree an open problem. This may explain why, still now, the percentage of data downloaded by RS stakeholders from the European Space Agency EO databases is estimated at about 10% or less [128].

ACKNOWLEDGMENTS

This work was supported in part by the National Aeronautics and Space Administration under Grant No. NNX07AV19G issued through the Earth Science Division of the Science Mission Directorate. The authors are very grateful to Prof. David W. Jacobs for helpful discussions and for revising the paper. A. Baraldi thanks Prof. Christopher Justice, Chair of the Department of Geographical Sciences at the University of Maryland, and Prof. Luigi Boschetti, currently at the Department of Forest, Rangeland and Fire Sciences, University of Idaho, for their support. The authors also wish to thank the Editor-in-Chief, Associate Editor and reviewers for their competence, patience and willingness to help.

REFERENCES

[1] M. Nagao and T. Matsuyama, *A Structural Analysis of Complex Aerial Photographs*. Plenum Press, New York, 1980.

[2] T. Matsuyama and V. S.-S. Hwang, *SIGMA: A Knowledge-Based Aerial Image Understanding System*. Plenum Press, New York, 1990.

[3] D. Marr, *Vision*. New York: Freeman and C., 1982.

[4] R. Capurro and B. Hjørland, "The concept of information," *Annual Review of Information Science and Technology*, vol. 37, pp. 343-411, 2003.

[5] N. Kumar, A. Berg, P. N. Belhumeur, and S. Nayar, "Describable visual attributes for face verification and image search," *IEEE Trans. Pattern Anal. Machine Intel.*, vol. 33, no. 10, pp. 1962–1977, 2011.

[6] R. Datta, D. Joshi, J. Li, and J. Z. Wang, "Image retrieval: Ideas, influences, and trends of the new age," *ACM Computing Surveys* (CSUR), vol. 40, no. 2, pp. 5-10, 2008.

[7] B. M. Mehtre, M. S. Kankanhalli, and W. F. Lee, "Shape measures for content based image retrieval: a comparison," *Info. Proc. Management*, vol. 33, no. 3, pp. 319–337, 1997.

[8] S. Jeannin (Ed.), MPEG-7 Visual part of experimentation model version 5.0, ISO/IEC JTC1/SC29/WG11/N3321, Nordwijkerhout, March, 2000.

[9] D. Zhang and G. Lu, "Evaluation of MPEG-7 shape descriptors against other shape descriptors," *Multimedia Systems*, vol. 9, pp. 15–30, 2003.

[10] T. Sikora, "The MPEG-7 Visual Standard for Content Description - An Overview," *IEEE Trans. Circuits Syst. Video Tech.*, vol. 11, no. 6, pp. 696-702, 2001.

[11] M. Bertero, T. Poggio, and V. Torre, "Ill-posed problems in early vision," *Proc. of the IEEE*, vol. 76, no. 8, pp. 869–889, Aug. 1988.

[12] *Perceptual Grouping*, Purdue University. [Online]. Available:
http://cs.iupui.edu/~tuceryan/research/ComputerVision/perceptual-grouping.html

[13] K. A. Stevens, Computation of locally parallel structure, *Bio. Cybernetics*, vol. 29, pp. 19-28, 1978.

[14] A. Baraldi and F. Parmiggiani, ``Combined detection of intensity and chromatic contours in color images,'' *Optical Engin.*, vol. 35, no. 5, pp. 1413-1439, May 1996.

[15] A. Baraldi and L. Boschetti, "Operational automatic remote sensing image understanding systems: Beyond geographic object-based and objectoriented image analysis (GEOBIA/GEOOIA). Part 1: Introduction," *Remote Sens.*, vol. 4, no. 9, pp. 2694–2735, Sep. 2012.

[16] A. Baraldi and L. Boschetti, "Operational automatic remote sensing image understanding systems: Beyond Geographic Object-Based and Object-Oriented Image Analysis (GEOBIA/GEOOIA) - Part 2: Novel system architecture, information/knowledge representation, algorithm design and implementation," *Remote Sens.*, vol. 4, pp. 2768-2817. 2012.

[17] C. Liedtke, J. Buckner, O. Grau, S. Growe, and R. Tonjes, "AIDA: A system for the knowledge based interpretation of remote sensing data," in *3rd Int. Airborne Remote Sensing* Conf., 1997.

[18] J. B¨uckner, M. Pahl, O. Stahlhut, and C. Liedtke, "geoAIDA – A knowledge based automatic image data analyser for remote sensing data," in ICSC Congress on Computational Intell. Methods Applic. (CIMA), 2001.

[19] A. Srivastava1, W. Mio, E. Klassen, and S. Joshi, "Geometric analysis of continuous, planar shapes," *Proc. 4th Int. Workshop on Energy Minimization Methods in Computer Vision and Pattern Recognition*, 2003.

[20] M. De Berg, O. Cheong, M. van Kreveld, and M. Overmars, Computational Geometry. Heidelberg, Germany: Springer, 2008.

[21] G. A. Miller, "The cognitive revolution: a historical perspective", in *Trends in Cognitive Sciences*, vol. 7, pp. 141-144, 2003.

[22] F. J. Varela, E. Thompson, and E. Rosch, The Embodied Mind: Cognitive Science and Human Experience. Cambridge, Mass.: MIT Press, 1991.

[23] D. Zhang and G. Lu, "Review of shape representation and description techniques," *Pattern Rec.*, vol. 37, no. 1, pp. 1–19, 2004.

[24] R. Laurini and D. Thompson, *Fundamentals of Spatial Information Systems*. London, UK: Academic Press, 1992.

[25] S. Skiena, The Algorithm Design Manual. London, UK: Springer-Verlag, 2008

[26] G. T. Toussaint, "Computational geometry and morphology," in Proc. 1st Int. Symposium for Science on Form, Tokyo, 1986, pp. 395-403.

[27] The Computational Geometry Algorithms Library (CGAL) [Online] Available: http://www.cgal.org/

[28] Library of Efficient Data Types and Algorithms (LEDA) [Online] Available: http://www.algorithmic-solutions.com/leda/about/changes_archiv.htm

[29] P. Soille, Morphological Image Analysis, Berlin: Germany: Springer-Verlag, 2003.

[30] *eCognition® Developer 9.0 Reference Book*, Trimble, 2015.

[31] *Open Source Computer Vision Library* (OpenCV). [Online]. Available: http://opencv.org/.

[32] *ENVI EX User Guide 5.0*, ITT Visual Information Solutions, Dec. 2009. [Online]. Available: http://www.exelisvis.com/portals/0/pdfs/enviex/ENVI_EX_User_Guide.pdf

[33] A Quality Assurance Framework for Earth Observation, version 4.0, Group on Earth Observation / Committee on Earth Observation Satellites (GEO/CEOS), 2010. [Online]. Available: http://qa4eo.org/docs/QA4EO_Principles_v4.0.pdf

[34] M. Page-Jones, *The Practical Guide to Structured Systems Design*. Englewood Cliffs, NJ, USA: Prentice-Hall, 1988.

[35] M. Peura and J. Iivarinen, "Efficiency of simple shape descriptors," in Proc. 3rd Int. Workshop on Visual Form, 1997,pp. 443–451.

[36] Si Liu, Hairong Liu, L. J. Latecki, Shuicheng Yan, Changsheng Xu, Hanqing Lu, "Size adaptive selection of most informative features," *Assoc. Advanc. Artificial Intel.*, 2011.

[37] H. Peng, H, F. Long, and C. Ding, "Feature selection based on mutual information: Criteria of max-dependency, max-relevance, and min-redundancy," *IEEE Trans. Pattern Anal. Machine Intell.*, vol. 27, pp. 1226–1238, 2005.

[38] C. M. Bishop, *Neural Networks for Pattern Recognition*. Oxford, U.K.: Clarendon, 1995.

[39] V. Cherkassky and F. Mulier, *Learning From Data: Concepts, Theory, and Methods*. Hoboken, NJ, USA: Wiley, 1998.

[40] T. Blaschke, G. J. Hay, M. Kelly, S. Lang, P. Hofmann, E. Addink, R. Queiroz Feitosa, F. van der Meer, H. van der Werff, F. van Coillie, and D. Tiede, "Geographic object-based image analysis - towards a new paradigm," *ISPRS J. Photogram. Remote Sens.*, vol. 87, pp. 180–191, Jan. 2014.

[41] L. A. Zadeh, "Fuzzy sets," *Inform. Control*, vol. 8, pp. 338–353, 1965.

[42] M. Sonka, V. Hlavac, and R. Boyle, *Image Processing, Analysis and Machine Vision*. London, U.K.: Chapman & Hall, 1994.

[43] J. Žunic, J. Pantovic and P. L. Rosin, "Measuring linearity of planar curves." A. Fred and M. De Marsico (eds.), in *Pattern Recognition Applications and Methods, Advances in Intelligent Systems and Computing* 318, DOI 10.1007/978-3-319-12610-4_16.

[44] P. L. Rosin, "Measuring shape: ellipticity, rectangularity, and triangularity," *Machine Vision and Applications*, vol. 14, no. 3, pp. 172–184, Jul. 2003.

[45] P. Mather, *Computer Processing of Remotely-Sensed Images - An Introduction*. Chichester, U.K.: Wiley, 1994.

[46] J. Piaget, Genetic Epistemology, New York: Columbia University Press, 1970.

[47] D. Parisi, "La scienza cognitive tra intelligenza artificiale e vita artificiale," in *Neurosceinze e Scienze dell'Artificiale: Dal Neurone all'Intelligenza*, Bologna, Italy: Patron Editore, 1991.

[48] J. Pearl, *Causality: Models, Reasoning and Inference*. New York (NY): Cambridge University Press, 2009.

[49] *Chi-Square Test for Independence*. Available online: http://stattrek.com/chi-square-test/independence.aspx. Accessed on 27 Feb. 2015.

[50] K. L. Delucchi, "The use and misuse of Chi-Square: Lewis and Burke revisited," *Psychological Bulletin*, vol. 94, no. 1, pp. 166-176, 1983.

[51] E. Kreyszig, *Applied Mathematics*. Wiley Press, 1979.

[52] D. Sheskin, *Handbook of Parametric and Nonparametric Statistical Procedures*. Boca Raton, FL: Chapman & Hall/CRC, 2000.

[53] B. G. Tabachnick and L. S. Fidell, *Using Multivariate Statistics*, Sixth Edition. Harlow, England: Pearson, 2014.

[54] S. Belongie, J. Malik, and J. Puzicha, "Shape matching and object recognition using shape contexts," *IEEE Trans. Pattern Anal. Machine Intel.*, vol. 24, no. 4, pp. 509–522, 2002.

[55] S. Belongie, G. Mori, and J. Malik, "Matching with shape contexts," in *Statistics and Analysis of Shapes*. Berlin/Heidelberg, Germany: Springer, 2006, pp. 81–105.

[56] S. Liang, *Quantitative Remote Sensing of Land Surfaces*. Hoboken, NJ, USA: Wiley, 2004.

[57] *Open source computer vision library* (OpenCV), [Online]. Available: http://opencv.org/. Accessed on March 20, 2015.

[58] J. Hadamard, "Sur les probl`emes aux d´eriv´ees partielles et leur signification physique," *Princeton University Bulletin*, vol. 13, pp. 49–52, 1902.

[59] M.K. Hu, "Visual pattern recognition by moment invariants," *IRE Trans. Inf. Theory IT*, vol. 8, pp. 179–187, 1962.

[60] *Distribution Selection - Cautions Regarding Goodness-of-Fit Tests*, United States Environmental Protection Agency (EPA). [Online]. Available: www.epa.gov/scipoly/sap/meetings/1998/march/attach3.pdf. Accessed on 27 Feb. 2015.

[61] A. Baraldi, L. Bruzzone, and P. Blonda, "Quality assessment of classification and cluster maps without ground truth knowledge," *IEEE Trans. Geosci. Remote Sens.*, vol. 43, no. 4, pp. 857–873, 2005.

[62] Z. Wang, A. C. Bovik, H. R. Sheikh, and E. P. Simoncelli, "Image quality assessment: From error visibility to structural similarity," *IEEE Trans. Image Proc.*, vol. 13, no. 4, pp. 1-14, 2004.

[63] L. Wald, T. Ranchin, and M. Mangolini, "Fusion of satellite images of different spatial resolutions: Assessing the quality of resulting images", *Photogram. Eng. Remote Sens.*, vol. 63, no. 6, pp. 691-699, 1997.

[64] Z. Wang and A. C. Bovik, "A universal image quality index," *IEEE Signal Proc. Letters*, vol. 9, no. 3, pp. 81-84, March 2002.

[65] L. Alparone, S. Baronti, A. Garzelli, and F. Nencini, "A global quality measurement of pan-sharpened multispectral imagery," *IEEE Geosci. Remote Sens. Letters*, vol. 1, no. 4, pp. 313-317, Oct. 2004.

[66] L. Alparone, B. Aiazzi, S. Baronti, A. Garzelli, and F. Nencini, "A new method for MS+Pan image fusion assessment without reference," *IEEE Proc.*, 2006.

[67] *World Happiness Report, 2012/2013*, Columbia University, Canadian Institute for Advanced Research, London School of Economics.

[68] A. K. Shackelford and C. H. Davis, "A combined fuzzy pixel-based and object-based approach for classification of high-resolution multispectral data over urban areas," *IEEE Trans. Geosci. Remote Sens.*, vol. 41, no. 10, pp. 2354–2363, 2003.

[69] A. K. Shackelford, *Development of urban area geospatial information products from high resolution satellite imagery using advanced image analysis techniques*. Ph.D. dissertation, University of Missouri- Columbia, 2004.

[70] R. Montero, "State of the art of compactness and circularity measures," *International Mathematical Forum*, vol. 4, no. 27, pp. 1305 – 1335, 2009.

[71] A. M. Maceachren, "Compactness of geographic shape: Comparison and evaluation of measures," *Geografiska Annaler. Series B, Human Geography*, vol. 67, no. 1, pp. 53-67, 1985.

[72] Wenwen Li, M. F. Goodchild, and R. L. Church, "An efficient measure of compactness for 2D shapes and its application in regionalization problems," *Int. J. of Geographical Information Science*, vol. 27, no. 6, pp. 1227-1250, 2013.

[73] A. Rosenfeld, "Compact figures in digital pictures," *IEEE Trans. Systems, Man and Cyber.*, vol. SMC-4, no. 2, pp. 221–223, 1974.

[74] L. Prechelt, ``A quantitative study of experimental evaluations of neural network learning algorithms: Current research practice," *Neural Networks*, vol. 9, 1996.

[75] A. Baraldi, L. Durieux, D. Simonetti, G. Conchedda, F. Holecz, and P. Blonda, "Automatic spectral-rule-based preliminary classification of radiometrically calibrated SPOT-4/-5/IRS, AVHRR/MSG, AATSR,

IKONOS/QuickBird/OrbView/GeoEye, and DMC/SPOT-1/-2 imagery – Part I: System design and implementation," *IEEE Trans. Geosci. Remote Sens.*, vol. 48, no. 3, pp. 1299–1325, 2010.

[76] A. Baraldi, V. Puzzolo, P. Blonda, L. Bruzzone, and C. Tarantino, "Automatic spectral rule-based preliminary mapping of calibrated landsat TM and ETM+ images," *IEEE Trans. Geosci. Remote Sens.*, vol. 44, no. 9, pp. 2563–2586, 2006.

[77] A. Baraldi, L. Durieux, D. Simonetti, G. Conchedda, F. Holecz, and P. Blonda, "Automatic spectral rule-based preliminary classification of radiometrically calibrated SPOT-4/-5/IRS, AVHRR/MSG, AATSR, IKONOS/QuickBird/OrbView/GeoEye, and DMC/SPOT-1/-2 imagery – Part II: Classification accuracy assessment," *IEEE Trans. Geosci. Remote Sens.*, vol. 48, no. 3, pp. 1326–1354, 2010.

[78] N. Kumar, P. N. Belhumeur, A. Biswas, D. W. Jacobs, W. J. Kress, I. C. Lopez, and J. V. B. Soares, "Leafsnap: A computer vision system for automatic plant species identification," in *European Conf. Computer Vision* (ECCV), 2012, pp. 502–516.

[79] J. V. B. Soares and D. W. Jacobs, "Efficient segmentation of leaves in semi-controlled conditions," *Machine Vision and Applications*, vol. 24, no. 8, pp. 1623–1643, Nov. 2013.

[80] L. Boschetti, S.P. Flasse, and P.A. Brivio, "Analysis of the conflict between omission and commission in low spatial resolution dichotomic thematic products: The Pareto boundary," *Remote Sens. Environ.*, vol. 91, pp. 280–292, 2004.

[82] R, B. D'Agostino and M. A. Stephens (Eds.), *Goodness-of-Fit Techniques*. New York: Marcel Dekker, Inc. Welch, B. L., 1986.

[83] R. S. Lunetta and C. D. Elvidge, *Remote sensing and Change Detection: Environmental Monitoring Methods and Applications*. Chelsea, MI, USA: Ann Arbor Press, 1998.

[81] R. G. Congalton and K. Green, *Assessing the Accuracy of Remotely Sensed Data*, Lewis Publishers: Boca Raton, 1999.

[84] D. S. Arnon and J. P. Gieselmann, "A linear time algorithm for the minimum area rectangle enclosing a convex polygon," *Tech. Rep.*, Purdue University, 1983.

[85] M. Pesaresi and J. A. Benediktsson, "A new approach for the morphological segmentation of high-resolution satellite imagery," *IEEE Trans. Geosci. Remote Sens.*, vol. 39, no. 2, pp. 309–320, 2001.

[86] Y. Chen, "New approaches for calculating Moran's index of spatial autocorrelation," *PLoS ONE*, vol. 8, no. 7, p. e68336, Jul. 2013.

[87] G. A. Moore, "Automatic scanning and computer processes for the quantitative analysis of micrographs and equivalent subjects," in *Pictorial Pattern Recog.*, Washington, DC, 1968, pp. 275–326.

[88] R. C. Gonzalez and R. E. Woods, *Digital Image Processing*, 3rd ed. Pearson/Prentice Hall, 2008.

[89] A. A. Melkman, "On-line construction of the convex hull of a simple polyline," *Info. Proces. Letters*, vol. 25, pp. 11–12, 1987.

[90] S. Tor and A. Middleditch, "Convex decomposition of simple polygons," *ACM Trans. on Graphics* (TOG), vol. 3, no. 4, pp. 244–265, 1984.

[91] B. Braden, "The surveyor's area formula," The College Mathematics Journal, vol. 17, no. 4, pp. 326–337, 1986.

[92] P. S. Heckbert and M. Garland, "Survey of polygonal surface simplification algorithms," in *Multiresolution Surface Modeling Course Notes*. ACM SIGGRAPH, 1997.

[93] D. H. Douglas and T. K. Peucker, "Algorithms for the reduction of the number of points required to represent a digitized line or its caricature," *Cartographica*, vol. 10, no. 2, pp. 112-122, 1973.

[94] A. Baraldi, "Fuzzification of a crisp near-real-time operational automatic spectral-rule-based decision-tree preliminary classifier of multisource multispectral remotely sensed images," *IEEE Trans. Geosci. Remote Sens.*, vol. 49, no. 6, pp. 2113 - 2134, June 2011.

[95] C. E. Woodcock, A. H. Strahler, and D. L. B. Jupp, "The use of variograms in remote sensing: I. Scene models and simulated images," *Remote Sens. Environ.*, vol. 25, no. 3, pp. 323–348, Aug. 1988.

[96] C. E. Woodcock, A. H. Strahler, and D. L. B. Jupp, "The use of variograms in remote sensing: II. Real digital images," *Remote Sens. Environ.*, vol. 25, no. 3, pp. 349–379, Aug. 1988.

[97] G. M. Espindola, G. Camara, I. A. Reis, L. S. Bins, A. M. Monteiro, "Parameter selection for region-growing image segmentation algorithms using spatial autocorrelation", *Int. J. Remote Sens.*, vol. 27, no. 14, pp. 3035–3040, 2006.

[98] W. R.Tobler, "A computer movie simulating urban growth in the Detroit Region," *Economic Geography*, vol. 46, pp. 234–240, 1970.

[99] A. Balaguer, L. A. Ruiz, T. Hermosilla, and J. A. Recio, "Definition of a comprehensive set of texture semivariogram features and their evaluation for object-oriented image classification," *Computers & Geosciences*, vol. 36, no. 2, pp. 231–240, Feb. 2010.

[100] M. Armstrong, *Basic Linear Geostatistics*. Berlin/Heidelberg, Germany: Springer, 1998.

[101] P.A. Longley, M.F. Goodchild, D.J. Maguire D. W. Rhind, *Geographic Information Systems and Science*, Second Edition. New York: Wiley, 2005.

[102] A. J. Lembo Jr., Spatial Autocorrelation:Moran's I and Geary's C, Salisbury University, 2000. [Online] Available: faculty.salisbury.edu/~ajlembo/419/sa.pdf

[103] L. Vincent, "Morphological reconstruction in image analysis: Applications and efficient algorithms," *IEEE Trans. Image Proc.*, vol. 2, no. 2, pp. 176–201, 1993.

[104] H. R. Wilson and J. R. Bergen, "A four mechanism model for threshold spatial vision," *Vis. Res.*, vol. 19, no. 1, pp. 19–32, 1979.

[105] J. Gil and M. Werman, "Computing 2D min, median, and max filters," *IEEE Trans. Pattern Anal. Machine Intel.*, vol. 15, no. 5, pp. 504–507, 1993.

[106] K. Robinson and P. F. Whelan, "Efficient morphological reconstruction: a downhill filter," *Pattern Recog. Let.*, vol. 25, no. 15, pp. 1759–1767, 2004.

[107] R. Adams, "Radial decomposition of disks and spheres," CVGIP: *Graphical Models and Image Proc.*, vol. 55, no. 5, pp. 325–332, 1993.

[108] L. Vincent, "Morphological grayscale reconstruction: definition, efficient algorithm and applications in image analysis," in *Computer Vision and Pattern Recog.* (CVPR), 1992, pp. 633–635.

[109] L. Lam, S.-W. Lee, and C. Y. Suen, "Thinning methodologies – a comprehensive survey," *IEEE Trans. Pattern Anal. Machine Intel.*, vol. 14, no. 9, pp. 869–885, 1992.

[110] M. Couprie, D. Coeurjolly, and R. Zrour, "Discrete bisector function and Euclidean skeleton in 2D and 3D," *Image and Vision Comput.*, vol. 25, no. 10, pp. 1543–1556, 2007.

[111] J. M. Cychosz, "Efficient binary image thinning using neighborhood maps," in *Graphics gems IV*, P. S. Heckbert, Ed. Academic Press Professional, Inc., 1994, pp. 465–473.

[112] A. Rosenfeld, "A characterization of parallel thinning algorithms," *Information and Control*, vol. 29, no. 3, pp. 286–291, 1975.

[113] A. Rosenfeld and A. C. Kak, *Digital Picture Processing*. Academic Press, Inc., 1982.

[114] M. A. Butt and P. Maragos, "Optimum design of Chamfer distance transforms", *IEEE Trans. Image Proc.*, vol. 7, no. 10, pp. 1477-1484, 1998.

[115] A. Meijster, J. B. T. M. Roerdink and W. H. Hesselink. "A general algorithm for computing distance transforms in linear time," in *Mathematical Morphology and its Applications to Image and Signal Processing*, Berlin/Heidelberg, Germany: Springer, 2000, pp. 331-340.

[116] M. Bertamini, "Who owns the contour of a visual hole?" *Perception*, vol. 35, pp. 883–894, 2006.

[117] J. V. B. Soares, A. Baraldi, and D. W. Jacobs, "Segment-based simple-connectivity measure design and implementation," *Tech. Rep.*, University of Maryland, College Park, 2014. [Online]. Available: http://hdl.handle.net/1903/15430.

[118] A. Soffer and H. Samet, "Negative shape features for image databases consisting of geographic symbols," in *Proc. 3rd Int. Workshop on Visual Form*, 1997.

[119] E. A. Wentz, "A shape definition for geographic applications based on edge, elongation, and perforation," *Geographical Anal.*, vol. 32, no. 2, pp. 95–112, 2000.

[120] G. Portoso, "On the maximum chi-square range of variation," *Dipartimento Scienze Economiche e Metodi Quantitativi*, Univ. degli Studi del Piemonte Orientale, 2008. [Online]. Available: http://semeq.unipmn.it/files/0814%20-%20Quaderno%20-%20Portoso.pdf. Accessed on: March 13, 2015.

[121] L. Hatcher, *Step-by-Step Basic Statistics Using SAS: Student Guide*. SAS Institute, 2003.

[122] G. Schaepman-Strub, M. E. Schaepman, T. H. Painter, S. Dangel, and J. V. Martonchik, "Reflectance quantities in optical remote sensing - Definitions and case studies," *Remote Sens. Environ*., vol. 103, pp. 27–42, 2006.

[123] Hanqiu Xu, "Modification of normalised difference water index (NDWI) to enhance open water features in remotely sensed imagery," Int. J. Remote Sens., vol. 27, no. 14, pp. 3025-3033, 2006.

[124] A. Baraldi, L. Boschetti, L., and M. Humber, "Probability sampling protocol for thematic and spatial quality assessments of classification maps generated from spaceborne/airborne very high resolution images," *IEEE Trans. Geosci. Remote Sens*., vol. 52, no. 1, Part: 2, pp. 701-760, Jan. 2014.

[125] A. Hamedianfara and H. Z. Mohd Shafr, "Detailed intra-urban mapping through transferable OBIA rule sets using WorldView-2 very-high-resolution satellite images," *Int. J. Remote Sens*., pp. 3380-3396, 2015.

[126] C. Shannon, "A mathematical theory of communication," *Bell System Technical Journal*, vol. 27, pp. 379–423 and 623–656, 1948.

[127] Group on Earth Observations, *GEO Announces Free and Unrestricted Access to Full Landsat Archive*. [Online] Available: www.fabricadebani.ro/ userfiles/GEO_press_release.doc

[128] S. D'Elia, European Space Agency, personal communication, 2012.

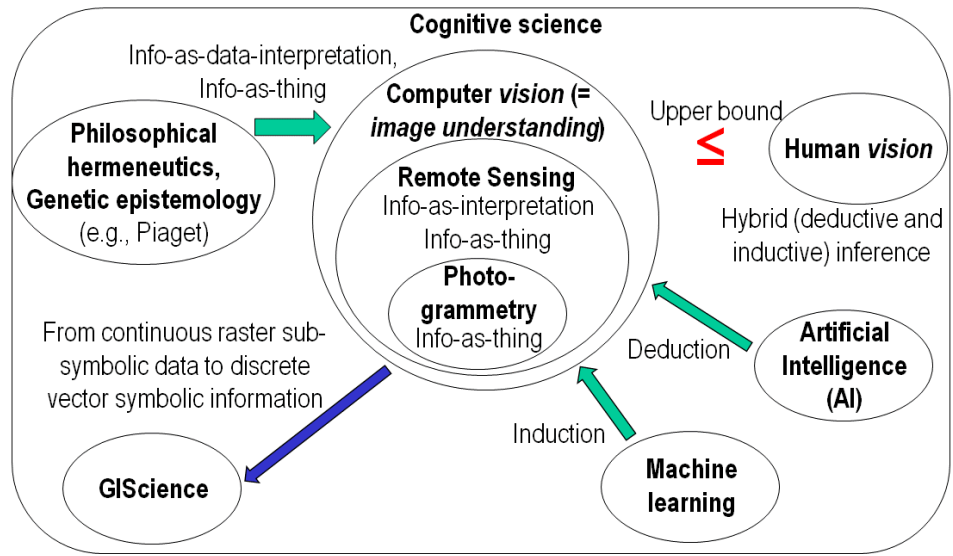

Fig. 1. Like engineering, remote sensing (RS) is a metascience, whose goal is to transform knowledge of the world, provided by other scientific disciplines, into useful user- and context-dependent solutions in the world [20]. Cognitive science is the interdisciplinary scientific study of the mind and its processes. It examines what cognition (learning) is, what it does and how it works. It especially focuses on how information/knowledge is represented, acquired, processed and transferred within nervous systems (humans or other animals) and machines (e.g., computers) [21], [22].

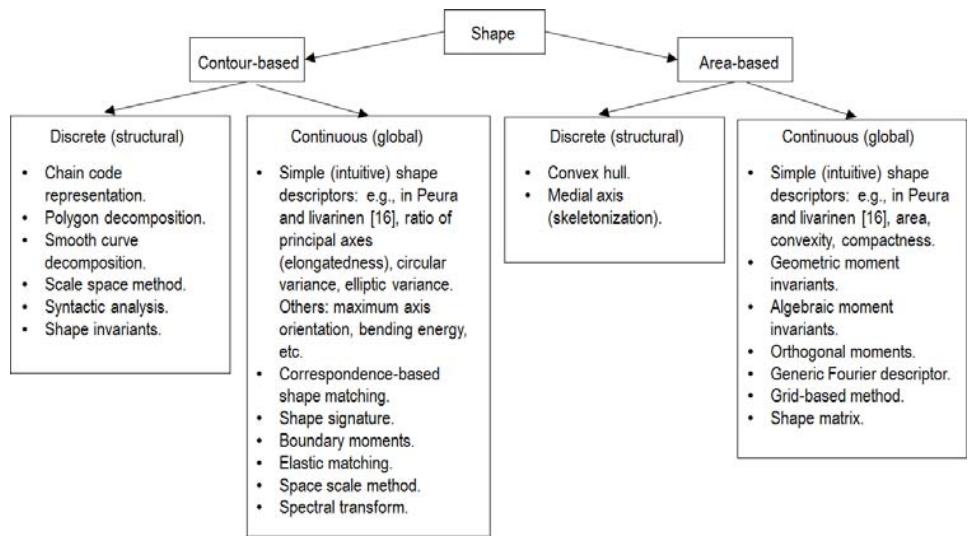

Fig. 2. Taxonomy of 2D/3D shape representation (modeling) and description (estimation) techniques, according to Zhang and Lu [9], [23]. Continuous approaches (global) do not divide shape into sub-parts. In the resulting multi-dimensional shape space [19], different shapes correspond to different points in the shape space and a quantification of shapes differences is accomplished using metrics between the acquired feature vectors based on geodesic lengths, e.g., Hamming distance, Hausdorff distance, comparing skeletons and support vector machine pattern matching, etc. Discrete approaches (structural) break the shape boundary into segments, called primitives using a particular criterion. The final representation is usually a string or a graph (or tree) and the similarity measure is accomplished by string matching or graph matching.

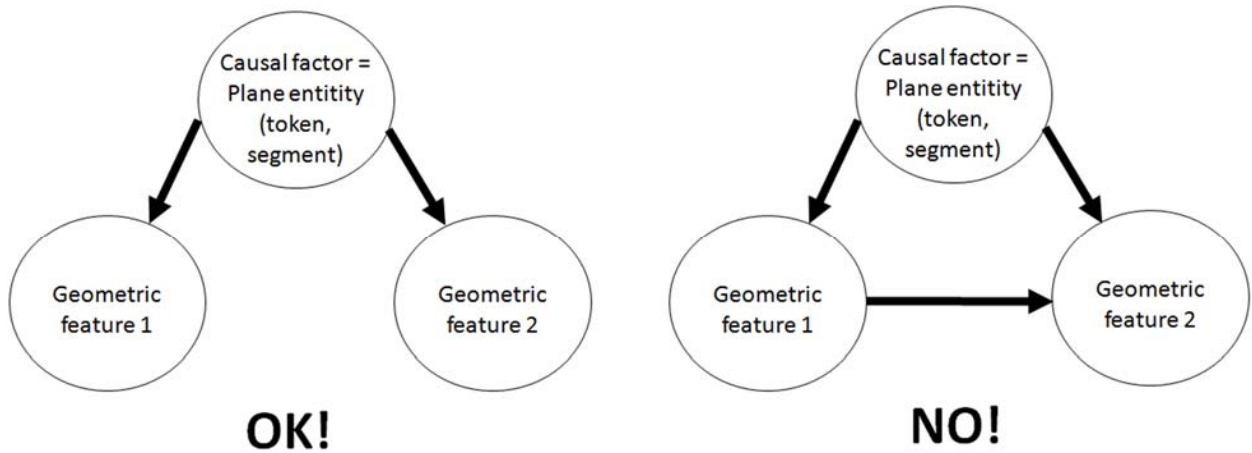

Fig. 3. A causal model is a directed acyclic graph G, which has no loops in it. Each vertex (node) has an associated random variable. All the arrows in a causal model indicate the possibility of a direct causal influence. All root vertices (with no parents) in the graph are labeled by independent random variables. No causal relationship between two geometric features is accepted [48].

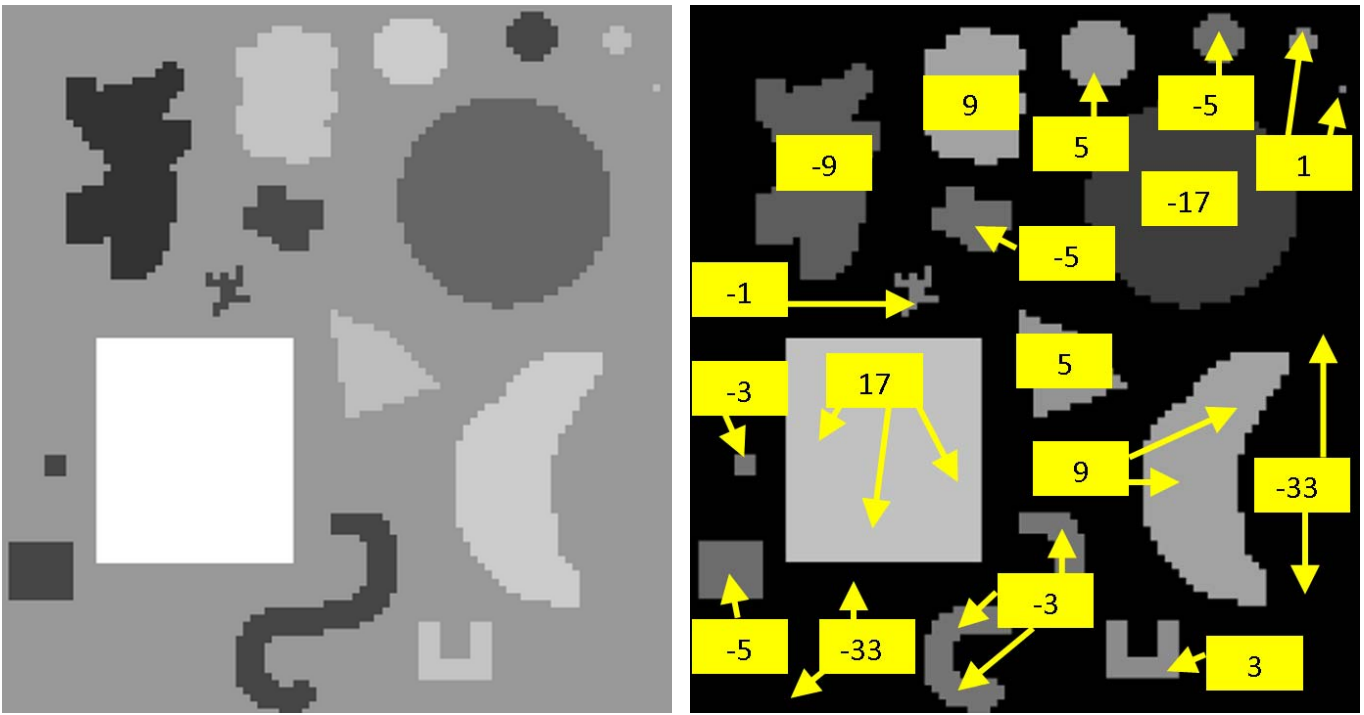

Fig. 4. (a) Left: Subset of a synthetic grayscale test image of known complexity, featuring several planar shapes of different size, selected to test the degree of match of the morphological multiscale characteristic with known object properties; (b) Right: Per-pixel morphological multiscale characteristic values. A negative (vice versa, positive) value indicates that the closing (vice versa, opening) differential morphological profile (DMP) contains the largest DMP response [85]. This occurs when the segment that pixel belongs to is darker (vice versa, lighter) than its background. The per-pixel characteristic scale is defined as the scale at which the DMP response is maximum. The characteristic scale thus provides an estimate of the local size (in pixel unit) of the image-object that pixel belongs to. A segment-based average of the per-pixel characteristic scale is an estimate of the size of the image-object. It is shown in yellow highlight. In this experiment, the adopted dyadic scales were s = 0, 1, 3, 5, 9, 17, 33 in pixel unit (see text).

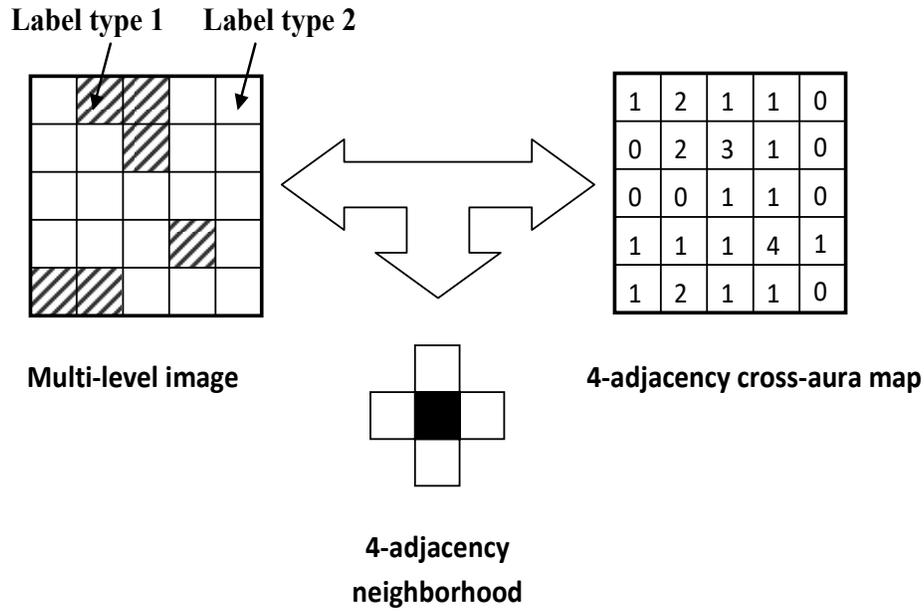

Fig. 5. Four-adjacency cross-aura map computed from a multi-level image. Every pixel value in the four-adjacency cross-aura map is equivalent to the number of four-adjacency neighboring pixels that do not belong to the same label type assigned to the central pixel.

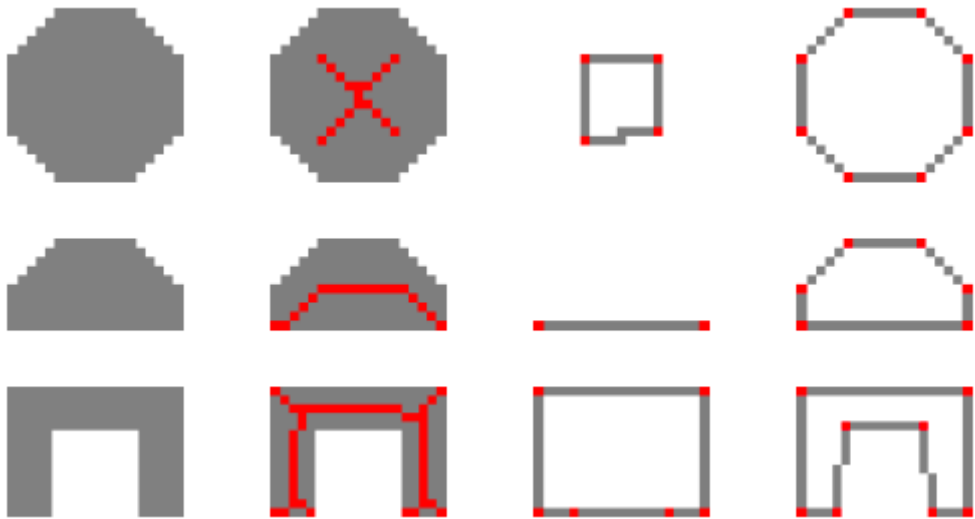

Fig. 6. Regions where polygon approximation by skeleton endpoints fails. From left to right, columns represent: (i) original region, (ii) skeleton (depicted in red) obtained by region thinning, (iii) polygonal approximation from skeleton endpoints (depicted in red), (iv) polygonal approximation using the Ramer-Douglas-Peucker (RDP) algorithm [92], [93]. In column (iii), the skeleton-based polygonal approximation shows problems in the first and second rows due to the rounded region boundaries where no skeleton endpoint is located. The third row appears affected by a similar effect due to concavities.

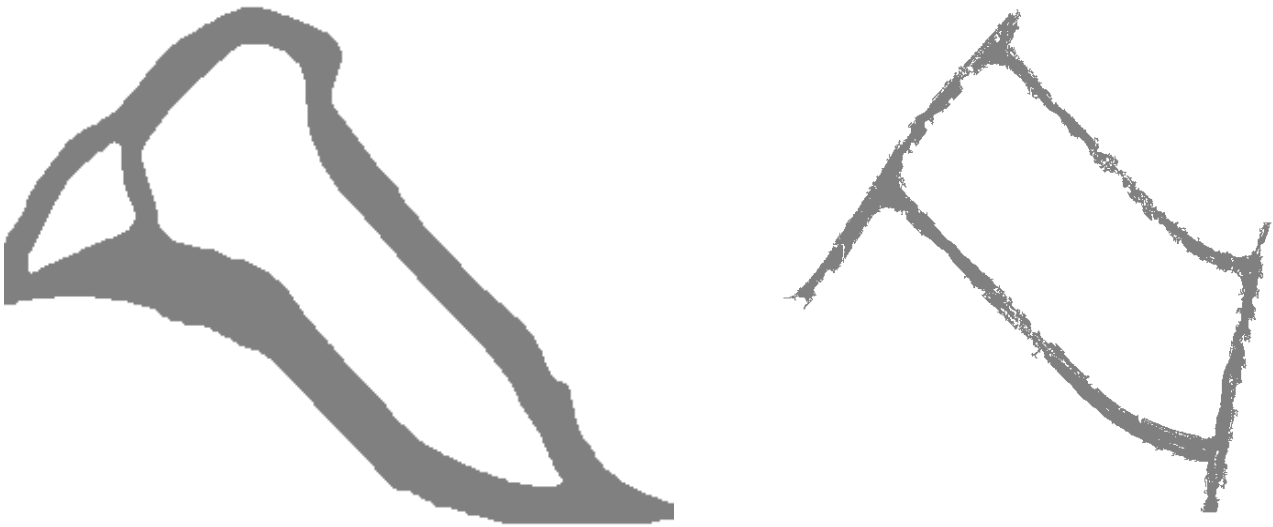

Fig. 7. Spaceborne VHR test image. (a) Left: A segment of a river, containing large holes. (b) Right: A segment of a road, containing large and small holes. If these holes were filled in, these segments would score low in *elongatedness*, whereas human experts would expect segments of rivers and roads to score high in *elongatedness*.

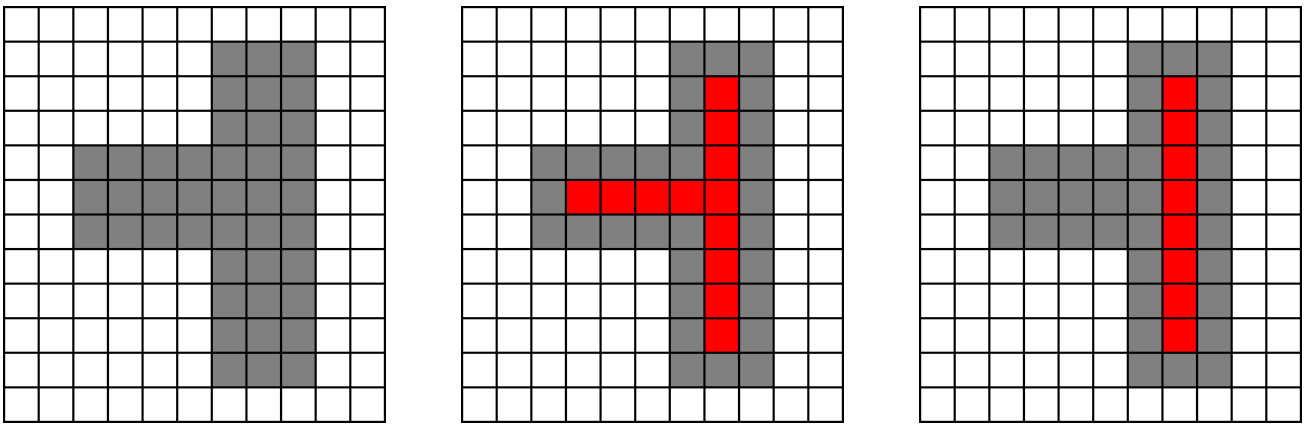

Fig. 8. (**a**) Left: Test plane object number 1. (**b**) Center: Skeleton of the test region. (**c**) Right: The longest path along the skeleton, estimated according to Nagao & Matsuyama [1]. This test illustrates how popular alternative formulations of elongatedness perform differently. For example, there is a large difference in results if the region's elongatedness is estimated using either the whole skeleton, like in Equation (18), where L = 12 pixels, or just its longest path, like in Equation (17), where $L_{NM}$ = 8. Although different methods compute local widths differently, for the purposes of illustration, let us assume the width computed at every point on the skeleton is equal to 3 pixels. Therefore, W = 3 in Equation (18) and $W_{NM}$ = 3 in Equation (17). Hence, *ElngtdnssAndNoHole* = Equation (18) = L / W = 12/3 = 4, whereas *Elngtdnss$_{NM}$* = Equation (17) = $L_{NM}$ / $W_{NM}$ = 8/3 = 2.67.

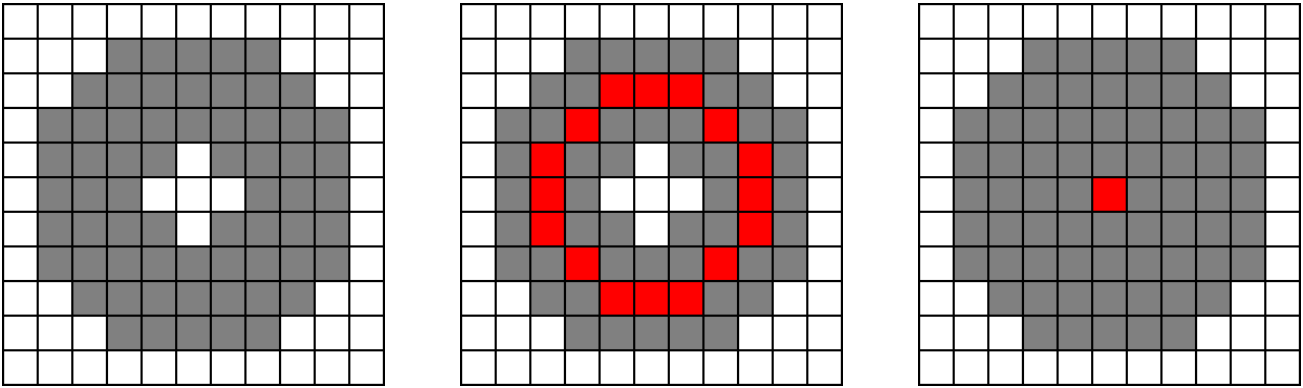

Fig. 9. (**a**) Left: Test plane object number 2. (**b**) Center: Skeleton of the test region. (**c**) Right: Skeleton of the test region whose hole was filled in, as required in [1]. When the hole is not filled in, the length value of the skeleton is larger, while the segment's average width is smaller. The elongatedness measure we propose here uses the skeleton depicted in the center, so that *ElngtdnssAndNoHole* = L / W = 16.0 / 4.1 = 3.9. On the other hand, the method by Nagao & Matsuyama [1] will use the skeleton depicted on the right, resulting in *Elngtdnss$_{NM}$* = L$_{NM}$ / W$_{NM}$ = 1 / 10 = 0.1.

| Segment | FilledAreaRatio | Simple-Connectivity 4-Adjacency | Simple-Connectivity = min{FilledAreaRatio, Simple-Connectivity 4-Adjacency} |
|---------|-----------------|----------------------------------|------------------------------------------------------------------------------|
| Boundary of holes increases | 0.84 | 0.71 | 0.71 |
| Area of holes increases | 0.84 | 0.56 | 0.56 |
| | 0.31 | 0.55 | 0.31 |

Fig. 10. Examples of simple connectivity measures. Since the total area of the holes in the top and middle segments shown in the left column is the same, then their *FilledAreaRatio* term is the same (equal to 0.84). However, in these two segments the spatial distribution of holes and the inner boundary length is different. The *SmplCnctvty4Adjcncy* term captures these differences. In the middle and bottom segments of the left column, the boundary lengths are similar, but the total areas of the holes are different. In this case, the *FilledAreaRatio* term is able to capture this difference (presenting values, respectively, of 0.84 and 0.31), while the *SmplCnctvty4Adjcncy* is not. The final measure, *CombndSmplCnctvty*, where a fuzzy-AND (minimum) operator combines the two membership functions *FilledAreaRatio* and *SmplCnctvty4Adjcncy*, behaves in good agreement with human perception.

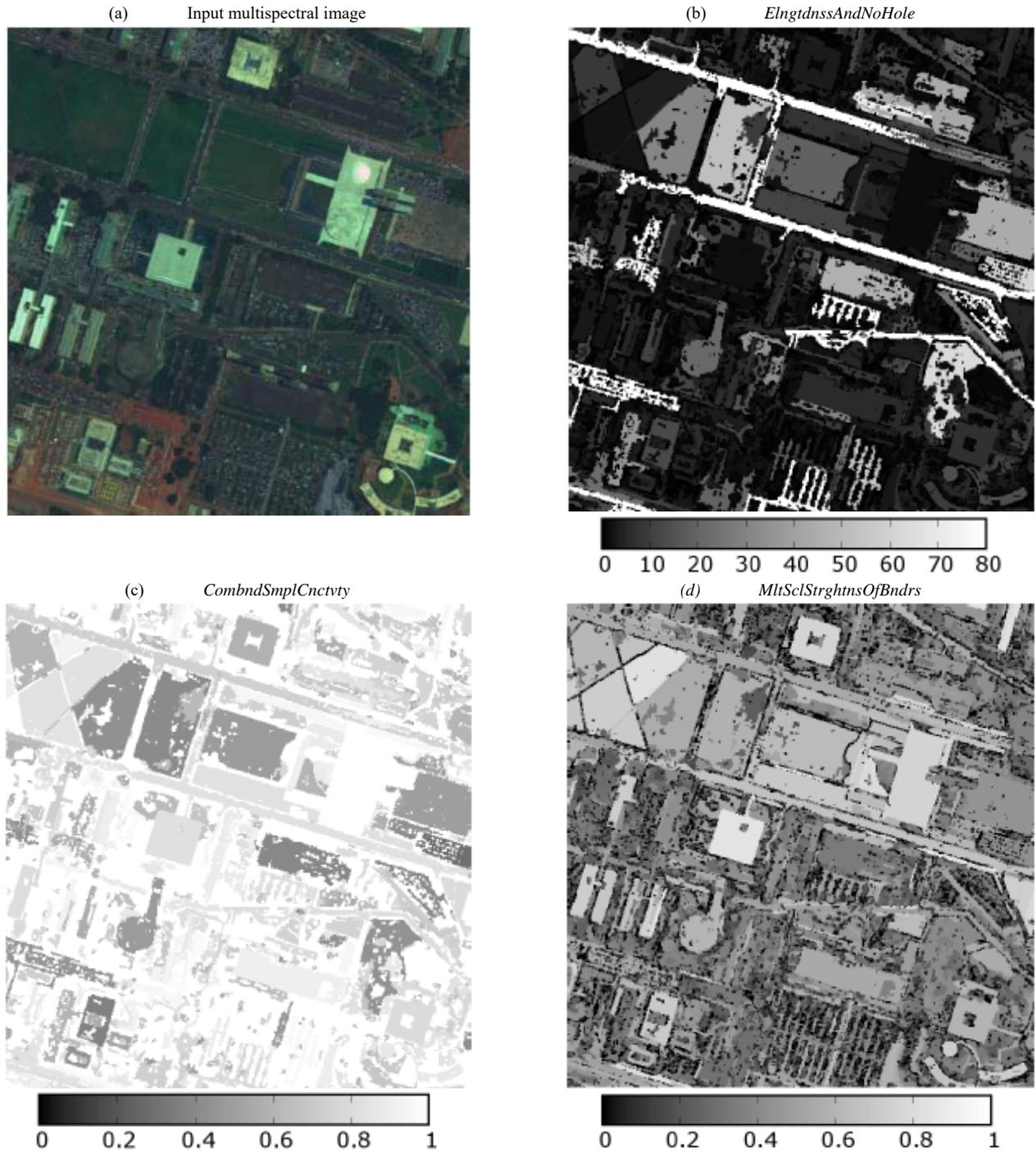

Fig. 11. (**a**) Zoom in of a WorldView-2 multispectral image of Brazilia, Brazil, depicted in false colors (R channel: Visible Red; G channel: Near Infrared; Blue channel: Visible Blue). Spatial resolution: 2 m. Acquisition date and time: 2010-08-04, 13:32 p.m.; (**b**) *ElngtdnssAndNoHole* planar feature values; (**c**) *CombndSmplCnctvty* planar feature values. When simple connectivity scores low, then area-based geometric indexes, including *ElngtdnssAndNoHole*, *CnvxtyAndNoHole* and *RndnssAndNoHole*, should be considered biased by the presence of holes; (**d**) *MltSclStrghtnsOfBndrs* planar feature values.

| Segment Number | Chromatic | Panchromatic | Segment | Convexity and No Hole | Elongatedness | Polygon-Based Approximate Rectangularity | Roundness and No Hole | Simple-Connectivity | Straightness of Boundary | Angle of MER (in degrees) | Area (in pixels) | Average Contrast Along Boundary | Morphological multiscale characteristic | Mean Panchromatic intensity |
|---|---|---|---|---|---|---|---|---|---|---|---|---|---|---|
| 1 | | | | 0.96 | 1.10 | 1.00 | 0.90 | 1.00 | 0.68 | 90.00 | 81 | 30.53 | 5.59 | 94.95 |
| 2 | | | | 0.85 | 2.92 | 0.95 | 0.66 | 1.00 | 0.63 | 75.07 | 666 | 1.91 | 15.14 | 57.62 |
| 3 | | | | 0.86 | 4.68 | 1.00 | 0.62 | 1.00 | 0.77 | -16.50 | 237 | 36.07 | 5.12 | 113.29 |
| 4 | | | | 0.87 | 8.72 | 1.00 | 0.53 | 0.72 | 0.83 | 74.05 | 1406 | 27.24 | 7.27 | 89.84 |
| 5 | | | | 0.78 | 4.86 | 1.00 | 0.58 | 0.89 | 0.89 | 73.30 | 1812 | 27.00 | 15.62 | 76.79 |
| 6 | | | | 0.48 | 9.24 | 0.78 | 0.44 | 1.00 | 0.79 | 155.85 | 461 | 7.83 | 16.31 | 59.71 |
| 7 | | | | 0.35 | 50.42 | 0.72 | 0.22 | 1.00 | 0.89 | -105.95 | 727 | 17.72 | 9.64 | 54.21 |
| 8 | | | | 0.67 | 22.35 | 0.05 | 0.33 | 1.00 | 0.85 | -5.57 | 340 | 20.51 | 8.87 | 55.27 |
| 9 | | | | 0.84 | 9.61 | 1.00 | 0.54 | 0.93 | 0.85 | 167.83 | 555 | 20.22 | 8.67 | 49.82 |

Fig. 12. Screenshot of the GUI specifically developed to show a human expert values of the proposed set of geometric attributes. In this GUI, darker cells correspond to: (i) higher values of geometric attributes and (ii) lower values of photometric attributes, like the panchromatic mean intensity shown at the rightest column. In this figure, for reasons of readability only nine segments are shown simultaneously for comparison. Detected in the spaceborne VHR test image of an urban area, segments 1 through 6 correspond to buildings or parts of buildings while segments 7 through 9 belong to roads. These two families of segments appear easy to discriminate based on different combinations of ranges of change of their geometric attributes.

| Segment Number | Chromatic | Panchromatic | Segment | Convexity and No Hole | Elongatedness | Polygon-Based Approximate Rectangularity | Roundness and No Hole | Simple-Connectivity | Straightness of Boundary | Angle of MER (in degrees) | Area (in pixels) | Average Contrast Along Boundary | Morphological multiscale characteristic | Mean Panchromatic intensity |
|---|---|---|---|---|---|---|---|---|---|---|---|---|---|---|
| 1 | 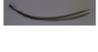 | 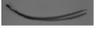 | 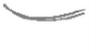 | 0.34 | 46.14 | 0.00 | 0.24 | 0.91 | 0.93 | -3.37 | 356 | 39.65 | 1.39 | 58.85 |
| 2 | 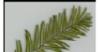 | 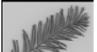 | 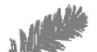 | 0.68 | 125.64 | 0.00 | 0.16 | 0.70 | 0.70 | -63.00 | 4502 | 32.39 | 3.05 | 109.80 |
| 3 | 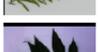 | 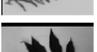 | 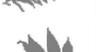 | 0.51 | 29.78 | 0.00 | 0.23 | 0.88 | 0.83 | 91.15 | 3444 | 42.87 | 7.49 | 40.39 |
| 4 | 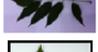 | 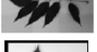 | 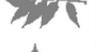 | 0.59 | 14.52 | 0.00 | 0.36 | 0.94 | 0.88 | 177.61 | 6004 | 59.83 | 15.34 | 68.36 |
| 5 | 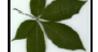 | 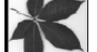 | 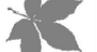 | 0.95 | 1.31 | 0.82 | 0.75 | 1.00 | 0.95 | 116.03 | 5112 | 43.14 | 30.41 | 43.30 |
| 6 | 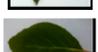 | 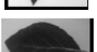 | 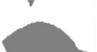 | 0.85 | 3.47 | 0.09 | 0.65 | 1.00 | 0.82 | 157.46 | 2765 | 32.60 | 29.69 | 32.74 |
| 7 | 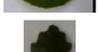 | 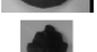 | 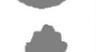 | 0.39 | 4.21 | 0.00 | 0.30 | 1.00 | 0.80 | -118.86 | 2462 | 45.73 | 7.61 | 74.82 |
| 8 | 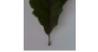 | 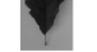 | 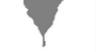 | 0.63 | 2.11 | 0.32 | 0.48 | 1.00 | 0.80 | 122.59 | 2692 | 45.92 | 14.87 | 25.25 |
| 9 | 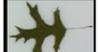 | 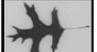 | 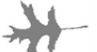 | 0.61 | 3.43 | 0.00 | 0.35 | 1.00 | 0.69 | 37.57 | 3927 | 42.51 | 28.30 | 49.09 |

Fig. 13. Screenshot of the GUI specifically developed to show a human expert values of the proposed set of geometric attributes. In this GUI, darker cells correspond to: (i) higher values of geometric attributes and (ii) lower values of photometric attributes, like the panchromatic mean intensity shown at the rightest column. In this figure, for reasons of readability only nine segments are shown simultaneously for comparison. Segments 1 through 9 are examples of segments extracted from pictures of leaves of different tree species, to be discriminated by different combinations of values of their geometric attributes.

TABLE CAPTIONS

| Two-way contingency table, Pearson's chi-square test for independence | | 1 | 2 | 3 | 4 | 5 | 6 | 7 |
|---|---|---|---|---|---|---|---|---|
| Feature ID | Feature Name | DMPmltSclChrctrstc | FuzzyRuleBsdRctnglrty (independent of holes) | RndnsAndNoHole | MltSclStrghtnsOfBndrs (independent of holes) | ElngtdnssAndNoHole | CombndSmplCnctvty | CnvxtyAndNoHole |
| 1 | DMPmltSclChrctrstc | - | 0.999946446495493 INDEP | 0.999998393392187 INDEP | 0.999991629040779 INDEP | 0.952786688564 2743 INDEP | 0.999981890258323 INDEP | 0.999999964292355 INDEP |
| 2 | FuzzyRuleBsdRctnglrty (independent of holes) | - | - | 0.000840119980309 INV RELATED | 0.995384972279882 INDEP | 0.00001752479 85527093 INV RELATED | 0.921252608090429 INDEP | 0.698323184142 6978 INDEP |
| 3 | RndnsAndNoHole | - | - | - | 0.999997029144091 INDEP | 8.32795257318 881E-235 INV RELATED | 0.00058785970 797259 DIR RELATED | 1.08915774609 503E-83 DIR RELATED |
| 4 | MltSclStrghtnsOfBndrs (independent of holes) | - | - | - | - | 0.943657449426006 INDEP | 0.999998540333204 INDEP | 0.98554120880 859 INDEP |
| 5 | ElngtdnssAndNoHole | - | - | - | - | - | 2.17126001246 908E-25 INV RELATED | 4.92572204842 01E-36 INV RELATED |
| 6 | CombndSmplCnctvty | - | - | - | - | - | - | 0.70498998420 8181 INDEP |
| 7 | CnvxtyAndNoHole | - | - | - | - | - | - | - |

Table 1. Instances of the probability value (P-value) for the Pearson's chi-square test for independence. If the P-value for the chi-square test for independence, such that P-value = Probability(chi-square value > test chi-square value), is less than the adopted level of significance α = 0.05, then the null hypothesis H0 (the two random variables are independent) is rejected at (1 - α) = 95% level of confidence. Cells in gray deserve further investigation to avoid inter-feature causality.

| Two-way contingency table, Cramer's V index (CVI) | | 1 | 2 | 3 | 4 | 5 | 6 | 7 |
|---|---|---|---|---|---|---|---|---|
| Feature ID | Feature Name | DMPmltSclChrrctrstc | FuzzyRuleBsdRctnglrty (independent of holes) | RndnsAndNoHole | MltSclStrghtnsOfBndrs (independent of holes) | ElngtdnssAndNoHole | CombndSmplCnctvty | CnvxtyAndNoHole |
| 1 | DMPmltSclChrctrstc | - | 0.038739962314 3401 INDEP | 0.0422184875 974543 INDEP | 0.04274645555 95364 INDEP | 0.04641529951 86267 INDEP | 0.03835915618 11503 INDEP | 0.0399422366 815894 INDEP |
| 2 | FuzzyRuleBsdRctnglrty (independent of holes) | - | - | 0.0604386176 657904 INDEP | 0.04144600060 60964 INDEP | 0.04445032905 68624 INDEP | 0.02487057447 1763 INDEP | 0.0490802657 687802 INDEP |
| 3 | RndnsAndNoHole | - | - | - | 0.04211849797 60391 INDEP | 0.19759537992 4897 INDEP | 0.06175873540 69437 INDEP | 0.1076161185 68181 INDEP |
| 4 | MltSclStrghtnsOfBndrs (independent of holes) | - | - | - | - | 0.04680086977 3228 INDEP | 0.03740465401 82519 INDEP | 0.0458727712 064827 INDEP |
| 5 | ElngtdnssAndNoHole | - | - | - | - | - | 0.06908443779 571 INDEP | 0.0942004574 163843 INDEP |
| 6 | CombndSmplCnctvty | - | - | - | - | - | - | 0.0480834141 993999 INDEP |
| 7 | CnvxtyAndNoHole | - | - | - | - | - | - | - |

Table 2. Values of the Cramer's V index = CVI = Pearson's chi-square index (same as in Table 11-1) / Maximum of the Pearson's chi-square index, equal to [N * (min(row, column in the contingency table) – 1)] = Normalized chi-square test for independence ∈ [0, 1]. Intuitively, if CVI tends to zero, then statistical independence holds. No known cut-off value is adopted though, to consider the CVI as "low" for independence.

| Two-way contingency table, Cramer's V index (CVI) | | 1 | 2 | 3 | 4 | 5 | 6 | 7 |
|---|---|---|---|---|---|---|---|---|
| Feature ID | Feature Name | DMPmltSclChrctrstc | FuzzyRuleBsdRctnglrty (independent of holes) | RndnsAndNoHole | MltSclStrghtnsOfBndrs (independent of holes) | ElngtdnssAndNoHole | CombndSmplCnctvty | CnvxtyAndNoHole |
| 1 | DMPmltSclChrctrstc | - | -0.096794295 | -0.095800252 | 0.164210063 | 0.075797658 | -0.018788534 | -0.111445368 |
| 2 | FuzzyRuleBsdRctnglrty (independent of holes) | - | - | 0.344483351 | -0.013300685 | -0.27239941 | 0.061261282 | 0.28369982 |
| 3 | RndnsAndNoHole | - | - | - | -0.270656762 | -0.935916078 | 0.561271973 | 0.884509053 |
| 4 | MltSclStrghtnsOfBndrs (independent of holes) | - | - | - | - | 0.315012068 | -0.082310281 | -0.203619857 |
| 5 | ElngtdnssAndNoHole | - | - | - | - | - | -0.612848806 | -0.795742144 |
| 6 | CombndSmplCnctvty | - | - | - | - | - | - | 0.335519155 |
| 7 | CnvxtyAndNoHole | - | - | - | - | - | - | - |

Table 3. The Spearman's rank cross-correlation coefficient (SRCC) assesses how well the relationship between two ranked variables can be described using a monotonically increasing or decreasing function, even if their relationship is not linear, unlike the Pearson's cross-correlation coefficient, PCC. Traditionally, in absolute values, (i) a cross-correlation coefficient ≥ 0.80 represents strong agreement (cells in dark gray), (ii) between 0.40 and 0.80 describes moderate agreement (cells in light gray), and (iii) ≤ 0.40 represents poor agreement. Cells in dark gray deserve further investigation to avoid inter-feature causality.

| Attribute type | Attribute name | Description |
|---|---|---|
| Spatial | | Formulas for calculating COMPACTNESS, CONVEXITY, SOLIDITY, ROUNDNESS, and FORM FACTOR are from Russ, J. C. (2002). The Image Processing Handbook, Fourth Edition. Boca Raton, FL: CRC Press. |
| 1 | AREA | Total area of the polygon, minus the area of the holes. Values are in map units. |
| 2 | LENGTH | The combined length of all boundaries of the polygon, including the boundaries of the holes. This is different than the MAXAXISLEN attribute. Values are in map units. |
| 3 | COMPACTNESS | A shape measure that indicates the compactness of the polygon. A circle is the most compact shape with a value of 1 / pi. The compactness value of a square is 1 / 2(sqrt(pi)). COMPACT = Sqrt (4 * AREA / pi) / outer contour length |
| 4 | CONVEXITY | Polygons are either convex or concave. This attribute measures the convexity of the polygon. The convexity value for a convex polygon with no holes is 1.0, while the value for a concave polygon is less than 1.0. CONVEXITY = length of convex hull / LENGTH |
| 5 | SOLIDITY | A shape measure that compares the area of the polygon to the area of a convex hull surrounding the polygon. The solidity value for a convex polygon with no holes is 1.0, and the value for a concave polygon is less than 1.0. SOLIDITY = AREA / area of convex hull |
| 6 | ROUNDNESS | A shape measure that compares the area of the polygon to the square of the maximum diameter of the polygon. The "maximum diameter" is the length of the major axis of an oriented bounding box enclosing the polygon. The roundness value for a circle is 1, and the value for a square is 4 / pi. ROUNDNESS = 4 * (AREA) / (pi * MAXAXISLEN2) |
| 7 | FORM FACTOR | A shape measure that compares the area of the polygon to the square of the total perimeter. The form factor value of a circle is 1, and the value of a square is pi / 4. FORMFACTOR = 4 * pi * (AREA) / (total perimeter)2 |
| 8 | ELONGATION | A shape measure that indicates the ratio of the major axis of the polygon to the minor axis of the polygon. The major and minor axes are derived from an oriented bounding box containing the polygon. The elongation value for a square is 1.0, and the value for a rectangle is greater than 1.0. ELONGATION = MAXAXISLEN / MINAXISLEN |
| 9 | RECTANGULAR FIT | A shape measure that indicates how well the shape is described by a rectangle. This attribute compares the area of the polygon to the area of the oriented bounding box enclosing the polygon. The rectangular fit value for a rectangle is 1.0, and the value for a non-rectangular shape is less than 1.0. RECT_FIT = AREA / (MAXAXISLEN * MINAXISLEN) |
| 10 | MAIN DIRECTION | The angle subtended by the major axis of the polygon and the x-axis in degrees. The main direction value ranges from 0 to 180 degrees. 90 degrees is North/South, and 0 to 180 degrees is East/West. |
| 11 | MAJAXISLEN | The length of the major axis of an oriented bounding box enclosing the polygon. Values are map units of the pixel size. If the image is not georeferenced, then pixel units are reported. |
| 12 | MINAXISLEN | The length of the minor axis of an oriented bounding box enclosing the polygon. Values are map units of the pixel size. If the image is not georeferenced, then pixel units are reported. |
| 13 | NUMHOLES | The number of holes in the polygon. Integer value. |
| 14 | HOLESOLRAT | The ratio of the total area of the polygon to the area of the outer contour of the polygon. The hole solid ratio value for a polygon with no holes is 1.0. HOLESOLRAT = AREA / outer contour area |

Table 4. Geometric descriptors implemented in the ENVI EX 5.0 commercial software product.